\pdfoutput=1

\documentclass[11pt]{article}

\usepackage{ACL2023}

\usepackage{times}
\usepackage{latexsym}
\usepackage{booktabs}
\usepackage{CJKutf8}

\usepackage[T1]{fontenc}

\usepackage[utf8]{inputenc}

\usepackage{microtype}

\usepackage{inconsolata}
\usepackage{todonotes}
\usepackage{amsmath}
\usepackage{multirow}
\usepackage{CJKutf8}
\usepackage{enumitem}
\usepackage{color}
\usepackage{xcolor}
\usepackage{colortbl}
\usepackage{devanagari}

\title{Contextual Label Projection for Cross-Lingual Structured Prediction} 

\author{Tanmay Parekh$^{\dagger}$ \ \ \
I-Hung Hsu$^{\ddagger}$ \ \ \
Kuan-Hao Huang$^{\mathsection}$ \\
{\bf Kai-Wei Chang$^{\dagger}$ \ \ \
Nanyun Peng$^{\dagger}$} \\
$^{\dagger}$Computer Science Department, University of California, Los Angeles \\
$^{\ddagger}$Information Science Institute, University of Southern California \\
$^{\mathsection}$Department of Computer Science, University of Illinois Urbana-Champaign\\
\texttt{\{tparekh, kwchang, violetpeng\}@cs.ucla.edu} \\
\texttt{\{ihunghsu\}@isi.edu}, \ \ 
\texttt{\{khhuang\}@illinois.edu}
  }

\begin{document}
\maketitle

\newcommand{\mypar}[1]{\vspace{0.35em}\noindent\textbf{#1}}
\newcommand{\SideNote}[2]{\todo[color=#1,size=\small]{#2}} 

\newcommand{\ihung}[1]{\SideNote{blue!40}{#1 --i-hung}}
\newcommand{\kuanhao}[1]{\SideNote{green!40}{#1 --kuan-hao}}
\newcommand{\tanmay}[1]{\SideNote{orange!40}{#1 --tanmay}}
\newcommand{\kaiwei}[1]{\SideNote{brown!40}{#1 --kai-wei}}
\newcommand{\violet}[1]{\SideNote{purple!40}{#1 --violet}}

\newcommand{\modelName}[0]{CLaP}
\newcommand{\markerBaseline}[0]{Marker-based}

\newcommand{\zh}[1]{\begin{CJK}{UTF8}{gbsn}#1\end{CJK}}
\newcommand{\hi}[1]{\dn #1}

\begin{abstract}

Label projection, which involves obtaining translated labels and texts jointly, is essential for leveraging machine translation to facilitate cross-lingual transfer in structured prediction tasks.
Prior research exploring label projection often compromise translation accuracy by favoring simplified label translation or relying solely on word-level alignments. 
In this paper, we introduce a novel label projection approach, \modelName{}, which translates text to the target language and performs \textit{contextual translation} on the labels using the translated text as the context, ensuring better accuracy for the translated labels.
We leverage instruction-tuned language models with multilingual capabilities as our contextual translator, imposing the constraint of the presence of translated labels in the translated text via instructions.
We benchmark \modelName{} with other label projection techniques on zero-shot cross-lingual transfer across 39 languages on two representative structured prediction tasks --- event argument extraction (EAE) and named entity recognition (NER), showing over 2.4 F1 improvement for EAE and 1.4 F1 improvement for NER.
We further explore the applicability of \modelName{} on ten extremely low-resource languages to showcase its potential for cross-lingual structured prediction.

\end{abstract}

\section{Introduction}
\label{sec:introduction}

\begin{figure}[t]
    \centering
    \includegraphics[width=0.95\linewidth]{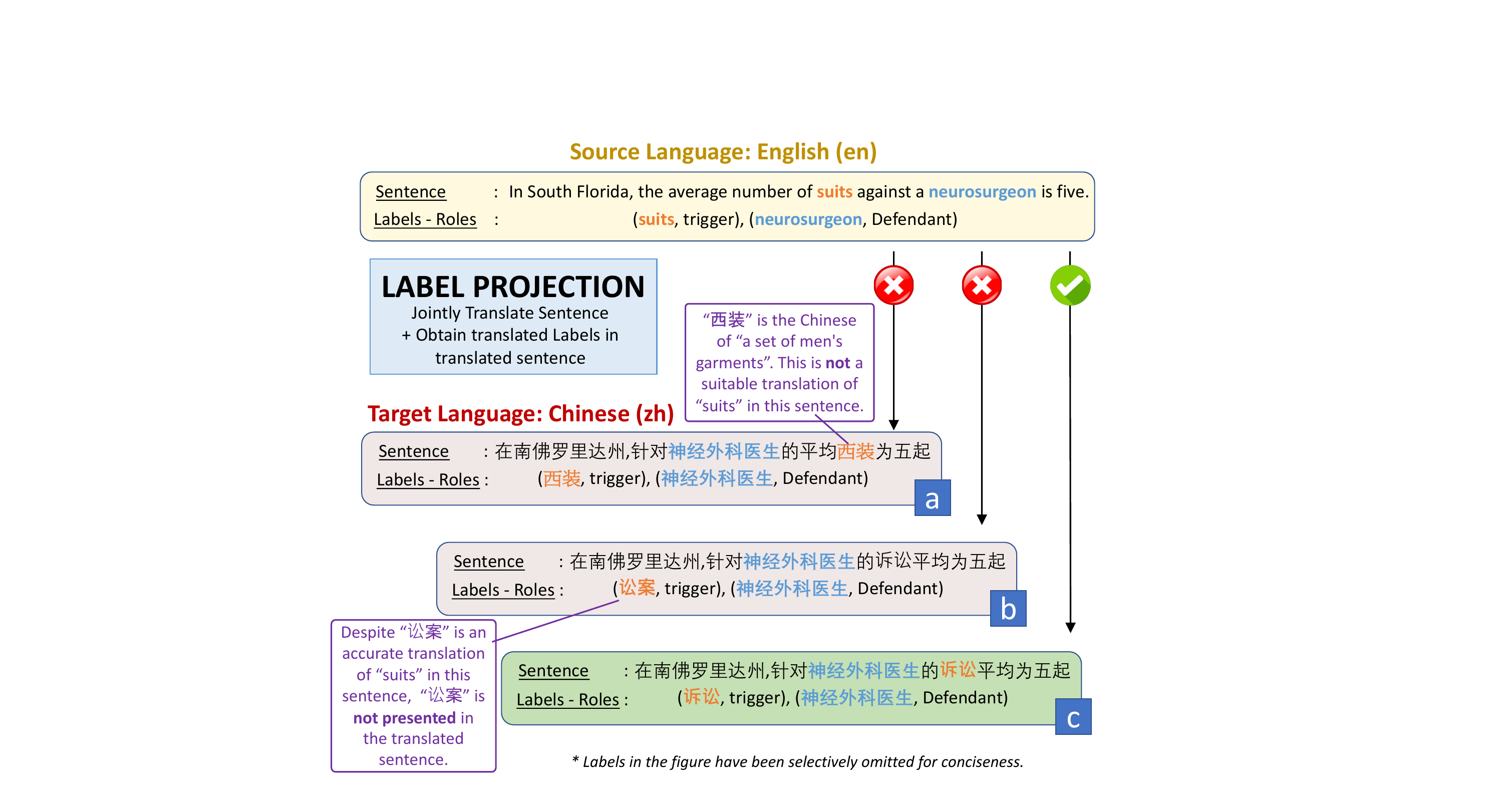}
    \caption{Illustration of the task of \textit{label projection} from English to Chinese.
    Label projection converts sentences from a source to a target language while translating the associated labels jointly. Failures in this process occur when (a) labels are either inaccurately translated or (b) missing in the translated sentence in the target language.}
    \label{fig:label-projection-task}
    \vspace{-0.9em}
\end{figure} 

Cross-lingual transfer for structured prediction tasks such as named entity recognition, relation extraction, and event extraction, has gained considerable attention recently~\cite{huang-etal-2022-multilingual-generative, DBLP:conf/aaai/CaoJ00023, tedeschi-navigli-2022-multinerd, huguet-cabot-etal-2023-red,  DBLP:conf/aaai/FinckeAMB22, jenkins-etal-2023-massively, DBLP:conf/aaai/AhmadPC21}.
It generalizes models trained in source languages to applications on other target languages~\cite{chen-ritter-2021-model, subburathinam-etal-2019-cross, pouran-ben-veyseh-etal-2022-mee}.

One effective and simple way to improve cross-lingual transfer performance 
is translate-train, which leverages machine translation techniques to generate pseudo-training data in the target languages by translating source language training data~\cite{xue-etal-2021-mt5, ruder-etal-2021-xtreme, yu-etal-2023-bridging}. 
However, adopting translate-train to structured prediction necessitates a \textit{label projection} step, which involves jointly translating input sentences and labels~\cite{chen-etal-2023-frustratingly}.
Label projection requires not only \textit{accurate translation} of the labels but also \textit{maintaining the association} between the translated texts and labels. As illustrated in Figure~\ref{fig:label-projection-task}, while ``\textit{suits}'' can have multiple valid translations, only ``\zh{诉讼}'' is present in the translated sentence and a proper translation at the same time.

\begin{figure*}[ht]
    \centering
    \includegraphics[width=0.95\linewidth]{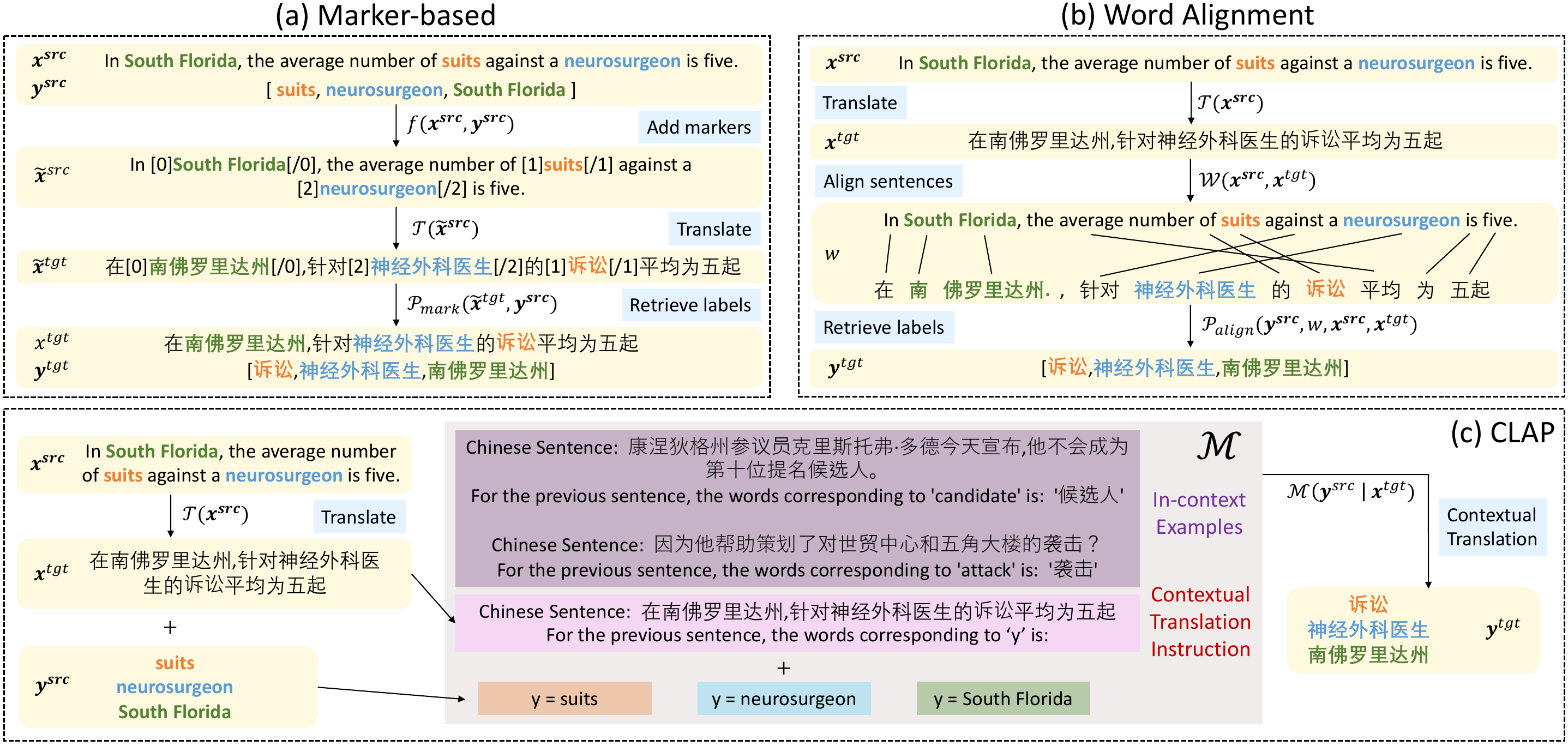}
    \caption{Illustration of the various techniques to conduct label projection: (a) \textbf{\markerBaseline{}} methods use markers to transform the sentence and translate the transformed sentence with label markers jointly, (b) \textbf{Word Alignment} methods use external word alignment tools to locate the translated labels in the translated sentence, and (c) \textbf{\modelName} (ours) performs contextual translation on labels using $\mathcal{M}$ (Here, we demonstrate the use of an instruction-tuned language model as $\mathcal{M}$ to identify translated labels within a translated sentence.).}
    \label{fig:methodology}
    \vspace{-0.8em}
\end{figure*}

Prior works have dealt with label projection through two primary frameworks.
The first one, illustrated in Figure~\ref{fig:methodology}(a), performs machine translation on modified source sentences that incorporate label annotations using special markers~\cite{chen-etal-2023-frustratingly,  hennig-etal-2023-multitacred}.
Translated labels can be extracted if special markers are retained in the translations.
In this approach, the quality of the translation is \textit{inherently compromised} due to the inclusion of special markers~\cite{chen-etal-2023-frustratingly}. 
The other framework uses word similarity to procure word alignments between the source and translated sentences.
Label translations are further constructed by combining mapped tokens in the translated sentence~\cite{stengel-eskin-etal-2019-discriminative, akbik-etal-2015-generating, aminian-etal-2019-cross}, as shown in Figure \ref{fig:methodology}(b).
However, it is hard for this framework to ensure \textit{accurate} label translation by merely using word alignments, as we will show in \S~\ref{sec:intrinsic-eval}.

In this work, we introduce \modelName{} (\textbf{C}ontextual \textbf{La}bel \textbf{P}rojection), which obtains projected label annotations by conducting contextual machine translation for the labels.
We first acquire the translation of the input sentence by any plug-and-play machine translator.
Then, inspired by the idea of contextual machine translation~\cite{wong-etal-2020-contextual, voita-etal-2018-context}, we use the translated input text as context to perform label translation, as shown in Figure~\ref{fig:methodology}(c).
Exploiting contextual machine translation strongly enhances the \textit{accuracy} of the translated labels while preserving their \textit{association} to the translated sentence.
Furthermore, translating the input sentence in an unmodified manner better leverages machine translators and assures the quality of translated sentence.
To implement contextual machine translation, we utilize a small instruction-tuned language model with multilingual capabilities, Llama-2-13B~\cite{DBLP:journals/corr/TouvronLlama23}.~\footnote{We also explore using GPT-3.5-Turbo in \S~\ref{sec:llm-clap}.}
We encode the translated input sentence and the constraint for the presence of labels in the form of instruction prompts and ask the language model to perform the label translation task.

Extensive experiments conducted on two representative tasks, event argument extraction (EAE) and named entity recognition (NER), reveal the following insights:
\begin{itemize}[topsep=2pt, itemsep=-2.5pt, leftmargin=13pt]
    \item Compared to existing label projection methods, \modelName{} performs the best on intrinsic evaluation by achieving the best label translation accuracy (\S~\ref{sec:intrinsic-eval}). Through extrinsic evaluation on downstream tasks, \modelName{} yields an average improvement of 2.4 and 1.4 F1 scores over the best baseline across 39 languages for EAE on ACE and NER on WikiANN datasets respectively (\S~\ref{sec:extrinsic-eval}).
    \item In comparison to directly prompting LLMs for the downstream task, we show that \modelName{}'s LLM usage for contextual machine translation provides significantly larger gains (\S~\ref{sec:extrinsic-eval}).
    \item Focusing on low-resource languages, \modelName{} demonstrates strong applicability by generalizing to ten extremely low-resourced African and American languages (\S~\ref{sec:low-resource}). Using larger LLMs for \modelName{} yields further improvements for low-resource languages, underlining \modelName{}'s future potential to improve continually (\S~\ref{sec:llm-clap}).
\end{itemize}
Our code can be found at \url{https:
//github.com/PlusLabNLP/CLaP}.

\section{Background}


\subsection{Structure Prediction Tasks}
\label{sec:str-pred}

Given an input sentence $\textbf{x}$, structure prediction models aim to predict structured output $\textbf{y}=[\textbf{x}[i_1:j_1], \textbf{x}[i_2:j_2], \dots, \textbf{x}[i_n:j_n]]$ (where $\textbf{x}[i_1:j_1]$ is an input sentence span from token $i_1$ to $j_1$) corresponding to a set of roles $\textbf{r}=[r_1, r_2, \dots, r_n]$ (where $r_i \in \mathcal{R}$, a pre-defined set of roles).
This vastly differs from standard classification-based tasks wherein the output prediction $y$ is a singular value from a fixed set of classes 
independent of the input sentence $\textbf{x}$.


\subsection{Zero-shot Cross-Lingual Transfer}
Zero-shot cross-lingual transfer~\cite{DBLP:conf/icml/HuRSNFJ20, ahmad-etal-2019-cross, huang-etal-2021-improving-zero} aims to train a downstream model for the target language $l_{tgt}$ using supervised data $\mathcal{D}_{src}$ from a source language $l_{src}$ without using any data in the target language (i.e. $\mathcal{D}_{tgt} = \phi$).
The paradigm has effectively advanced language technologies for under-resourced languages. 

\subsection{Translate-Train}
\label{sec:translate-train-background}

Translate-train \cite{DBLP:conf/icml/HuRSNFJ20, ruder-etal-2021-xtreme} is a popular and powerful zero-shot cross-lingual transfer technique that leverages machine translators $\mathcal{T}$ to boost downstream model performance.
Specifically, in translate-train, $\mathcal{D}_{src}$ is translated into the target language as pseudo training data $\mathcal{D}_{src}^{tgt}$ and the downstream model is trained using a combination of $\{\mathcal{D}_{src}, \mathcal{D}^{tgt}_{src}\}$.

Utilizing translate-train for structured prediction tasks 
requires \textit{Label Projection}, which includes two sets of translations: (1) Sentence translation ($\textbf{x}^{src} \rightarrow \textbf{x}^{tgt}$), where we use $\rightarrow$ to denote that $\textbf{x}^{tgt}$ is the transformation of $\textbf{x}^{src}$; and (2) Label translation ($\textbf{y}^{src} \rightarrow \textbf{y}^{tgt}$),
such that the translated label $\textbf{y}^{tgt}$ is appropriately \textit{associated with $\textbf{x}^{tgt}$}.
This demand makes translate-train for structure prediction tasks more complex than that for classification tasks, as the latter only requires sentence translation (since $y$ is independent of $\textbf{x}$).~\footnote{For certain structure prediction tasks like relation classification~\cite{DBLP:conf/aaai/AhmadPC21, hsu2021discourselevel} (determining the relationship between two entities in $\textbf{x}$), even if the output $y$ is scalar, translate-train necessitates label projection step due to the required projection of the two given entities into the translated sentence.}

\paragraph{Translate-Test}
Besides translate-train, translate-test is another commonly used technique in zero-shot cross-lingual transfer.
During inference, models trained on $\mathcal{D}_{src}$ are used to predict on translated test sentences ($\textbf{x}^{tgt} \rightarrow \textbf{x}^{src}$), and the predictions on $\textbf{x}^{src}$ are later mapped back to $\textbf{x}^{tgt}$. 
We mainly focus on translate-train in this work but discuss \modelName{}'s effectiveness for translate-test in \S~\ref{sec:translate-test}.

\subsection{Label Projection}
\label{subsec:label_project_definition}
We hereby technically define the problem of \textit{label projection} \cite{akbik-etal-2015-generating, chen-etal-2023-frustratingly}:
\begin{align*}
    & \textbf{x}^{src} \rightarrow \textbf{x}^{tgt} & \\
    \& \quad & y_{m}^{src} \rightarrow y^{tgt}_{m} & \quad \forall y_{m}^{src} \in \textbf{y}^{src} \\
    s.t. \quad & y^{tgt}_{m} \in \textbf{x}^{tgt} & \quad \forall y^{tgt}_{m} \in \textbf{y}^{tgt}.
\end{align*}
This problem requires optimizing two properties of accuracy and faithfulness in the translations:
\begin{itemize}[topsep=2pt, itemsep=-2.5pt, leftmargin=13pt]
    \item \textbf{Accuracy} ensures that $[\textbf{x}^{tgt}, y^{tgt}_{1}, \dots, y^{tgt}_{n}$] are accurate translations of $[\textbf{x}^{src}, y_{1}^{src}, \dots, y_{n}^{src}]$.
    \item \textbf{Faithfulness} ensures that each $y^{tgt}_{m}$ is associated with $\textbf{x}^{tgt}$ (the constraint of $y^{tgt}_{m} \in \textbf{x}^{tgt}$).
\end{itemize}
How to do this joint translation is non-trivial as standard translation models $\mathcal{T}$ cannot simply impose the additional faithfulness constraint, as shown in the failure cases in Figure~\ref{fig:label-projection-task}(b).
This demonstrates the challenge of label projection.

\section{Methodology}

In this section, we first formally define the previous attempts at label projection
and later introduce \modelName{}, which provides a new perspective of using contextual machine translation for label projection.

\subsection{Baseline Methods}
\label{subsec:baseline}

The primary frameworks used in prior works include \markerBaseline{} and Word-alignment methods.

\mypar{\markerBaseline{}} methods~\cite{lewis-etal-2020-mlqa, DBLP:conf/icml/HuRSNFJ20, chen-etal-2023-frustratingly}
solve the label projection by first marking labels to the input sentence $\textbf{x}^{src}$, forming $\tilde{\textbf{x}}^{src}$, and then use the translation model to obtain the potential translation of input sentence and labels jointly. For example, in Figure~\ref{fig:methodology}(a), ``South Florida'' is delineated by markers [0] and [\textbackslash0]. 
Assuming the preservation of markers after translation of $\tilde{\textbf{x}}^{src}$, a post-processing step, $\mathcal{P}_{mark}$, is performed to retain the translated labels $\textbf{y}^{tgt}$ and translated sentence $\textbf{x}^{tgt}$. Putting every step together, we have
\begin{align*}
    \tilde{\textbf{x}}^{src} = f(\textbf{x}^{src}, \textbf{y}^{src})&, \ \ \ \tilde{\textbf{x}}^{tgt} = \mathcal{T}(\tilde{\textbf{x}}^{src}) \\
    \textbf{x}^{tgt}, \textbf{y}^{tgt} &= \mathcal{P}_{mark}(\tilde{\textbf{x}}^{tgt}, \textbf{y}^{src}),
\end{align*}
where $f$ denotes the marker addition step and $\tilde{\textbf{x}}^{tgt}$ is the translation of $\tilde{\textbf{x}}^{src}$ using translator $\mathcal{T}$.

Despite their simplicity, these methods suffer from poor translation quality and reduced robustness to different translation models owing to their input sentence transformations and strong assumptions about the retention of markers in $\tilde{\textbf{x}}^{tgt}$.

\mypar{Word Alignment}
approaches \cite{akbik-etal-2015-generating, yarmohammadi-etal-2021-everything} first translate the input sentence and acquire word alignments \cite{dyer-etal-2013-simple, dou-neubig-2021-word} between the translation pairs. Each translated label $y_m^{tgt}$ is then procured by merging the aligned words of $y_m^{src}$ in the translated sentence using the word mappings $w$. For example, in Figure~\ref{fig:methodology}(b), the translated label for ``South Florida'' is obtained by merging two aligned words, which is done by a heuristic post-processing algorithm $\mathcal{P}_{align}$.
Formally, we have
\begin{align*}
    \textbf{x}^{tgt} = &\mathcal{T}(\textbf{x}^{src}), \ w = \mathcal{W}(\textbf{x}^{src}, \textbf{x}^{tgt}) \\
    y^{tgt}_{m} = &\mathcal{P}_{align}(y_{m}^{src}, w, \textbf{x}^{src}, \textbf{x}^{tgt}) & \forall y_{m}^{src} \in \textbf{y}^{src}
\end{align*}

Although these approaches deliver high-quality sentence translations, the accuracy of their translated labels is compromised. This is because the translated labels are reconstructed from word-level translations, lacking joint consideration of the entire span~\cite{akbik-etal-2015-generating, chen-etal-2023-frustratingly}.



\subsection{\modelName}


We tackle the task of label projection through a new perspective --- performing actual translation on labels instead of recovering them from translated text $\textbf{x}^{tgt}$.
This better ensures the accuracy of the translated labels $\textbf{y}^{tgt}$.
To accomplish this, we leverage the idea of \textit{contextual machine translation} on the label translation with $\textbf{x}^{tgt}$ as context.

Contextual machine translation, which aims to perform phrase-level translations conditional on the context of the translated sentence, is tangentially explored for applications like anaphora resolution \cite{voita-etal-2018-context} and pronoun translations \cite{wong-etal-2020-contextual}. 
The main goal of this task is to maintain the consistency of phrasal translations in the given context.
In our work, we develop a novel model \modelName{} to extend the idea of contextual translation to the application of label projection.

As illustrated in Figure~\ref{fig:methodology}(c),
\modelName{} first utilizes machine translation model $\mathcal{T}$ to translate input sentence $\textbf{x}^{src}$ to $\textbf{x}^{tgt}$.
Treating $\textbf{x}^{tgt}$ as the context, the contextual translation model $\mathcal{M}$ translates the labels $\textbf{y}^{src}$ to $\textbf{y}^{tgt}$.
Contextual translation implicitly imposes the \textit{faithfulness constraint} which requires $y^{tgt}_{m} \in \textbf{x}^{tgt} \quad, \forall y^{tgt}_{m} \in \textbf{y}^{tgt}$, hence, slackly satisfying the requirement of label projection.
These two steps can be formally described as:
\begin{align*}
    \textbf{x}^{tgt} &= \mathcal{T}(\textbf{x}^{src}) \\
    y^{tgt}_{m} &= \mathcal{M}(y_{m}^{src} | \textbf{x}^{tgt}) & \quad \quad \forall y_{m}^{src} \in \textbf{y}^{src}
\end{align*}
where $y^{tgt}_{m}$ is generated from $\mathcal{M}(y_{m}^{src} | \textbf{x}^{tgt})$, drawing the difference from the previous works.

Compared to word alignment approaches using simple word-similarity aligners $\mathcal{W}$, we use models with translation capabilities $\mathcal{M}$, to improve the accuracy of translated labels.
Furthermore, the independence of $\mathcal{T}$ and $\mathcal{M}$ for translating $\textbf{x}^{src}$ and $\textbf{y}^{src}$ respectively assures that \modelName{} has better translation quality for $\textbf{x}^{tgt}$ and is more robust than the marker-based baselines.
We empirically back these intuitions in \S~\ref{sec:intrinsic-eval}.

\subsection{Implementing \modelName{}}

To implement our concept, we first configure $\mathcal{T}$ to be a modular component that can be replaced by any third-party translation model.
For $\mathcal{M}$, we use an instruction-tuned language model (LM) with multilingual capabilities \cite{DBLP:journals/corr/abs-2109-01652, DBLP:journals/corr/abs-2211-05100}.
Instruction-tuned LMs can accept conditional information in their natural language prompt.
Specifically, we encode the translated target sentence $\textbf{x}^{tgt}$ as well as the faithfulness constraint $y^{tgt}_{m} \in \textbf{x}^{tgt}$ implicitly in the form of natural language instructions (highlighted as ``Contextual Translation Instruction'' in Figure~\ref{fig:methodology}(c)).
Following \citet{DBLP:journals/corr/abs-2005-14165}, we also provide $n$ randomly chosen in-context examples (highlighted as ``In-context examples'' in Figure~\ref{fig:methodology}(c)) to improve the instruction-understanding capability of the model.~\footnote{The in-context examples are generated using Google translation and initial prediction from instruction-tuned LMs. The label predictions are further verified by back-translation.}
Instruction-tuned LMs sacrifice some translation ability compared to supervised machine translation models~\cite{DBLP:journals/corr/abs-2304-04675}, however, they provide better control of contextual constraints.

After obtaining label translations, we employ simple string-matching algorithms to get the exact span index of $y^{tgt}_m$ in $\textbf{x}^{tgt}$. Though this may not be the optimal solution when duplicated strings exist in $\textbf{x}^{tgt}$, it works well in practice as stated in prior word-alignment methods \cite{dou-neubig-2021-word}.

\section{Experiments and Results}

This section outlines our experimental settings, which includes the datasets, baselines, and implementation details. Subsequently, we provide an in-depth analysis of \modelName{} through both intrinsic and extrinsic evaluations.

\subsection{Task and Dataset}

We choose two structure prediction tasks, event argument extraction (EAE) \cite{sundheim-1992-overview, hsu-etal-2023-tagprime} and named entity recognition (NER) \cite{tjong-kim-sang-2002-introduction, tjong-kim-sang-de-meulder-2003-introduction} for evaluating our label projection method.
EAE requires the extraction of text segments serving as arguments corresponding to an event and mapping them to their corresponding argument roles. NER aims to identify and categorize named entities from the input sentence.
For EAE, we use multilingual ACE dataset \cite{doddington-etal-2004-automatic} and follow the pre-processing by \citet{huang-etal-2022-multilingual-generative} to retain 33 event types and 22 argument roles.
For NER, we consider the WikiANN \cite{pan-etal-2017-cross, rahimi-etal-2019-massively} with pre-processing by \citet{DBLP:conf/icml/HuRSNFJ20}.
We list the basic statistics for these datasets in Table~\ref{tab:data-statistics} and more details in \S~\ref{sec:appendix-data-statistics}.
For experiment, we consider the zero-shot cross-lingual transfer using English (\textit{en}) as the source language.

\begin{table}[t!]
    \small
    \centering
    \begin{tabular}{lrr}
    \toprule
         & \textbf{ACE} & \textbf{WikiANN} \\
         \midrule
        \# Train Instances & 4,202 & 20,000 \\
        \# Dev Instances & 450 & 10,000 \\
        \# Avg. Test Instances & 194 & 6,469 \\
        \# Test Languages & 2 & 39 \\
    \bottomrule
    \end{tabular}
    \caption{High-level data statistics for ACE and WikiANN datasets for EAE and NER tasks respectively. \# = `number of' and Avg. = average.}
    \vspace{-0.5em}
    \label{tab:data-statistics}
\end{table}

\subsection{Baselines}
\label{sec:baseline_exp}
We select two label projection models as baselines, each representing the two baseline frameworks we covered in Section~\ref{subsec:baseline}, respectively:
(1) \textbf{EasyProject} \cite{chen-etal-2023-frustratingly}, a recent marker-based label-projection method, utilizes numbered square braces (e.g. [0] and [/0]) to mark the labels in the input sentence.
(2) \textbf{Awesome-Align} \cite{dou-neubig-2021-word}, a neural bilingual word alignment model, uses multilingual language models to find word similarities to derive word alignments, which are later used for label projection.

\subsection{Implementation Details}

For the translation model $\mathcal{T}$, we experiment with the Google Machine Translation (GMT)~\cite{DBLP:journals/corr/WuSCLNMKCGMKSJL16}.\footnote{\url{https://cloud.google.com/translate}. We use the free scraping tool to reduce the translation costs.}
For \modelName{}, we use the text-completion version of Llama-2 \cite{DBLP:journals/corr/TouvronLlama23} with 13B parameters 
as $\mathcal{M}$.
We use $n=2$ in-context examples for \modelName{} prompts.
For Awesome-align, we use the unsupervised version of their model utilizing multilingual BERT \cite{devlin-etal-2019-bert} as it provides better results \cite{chen-etal-2023-frustratingly}.~\footnote{We utilize the non fine-tuned version of EasyProject since we experiment using GMT. The original work also explores finetuning the machine translation model but it requires open-source access for finetuning.}
Additional details are provided in Appendix~\ref{sec:appendix-implementation-details}.

\begin{figure}
    \centering
    \includegraphics[width=0.47\textwidth]{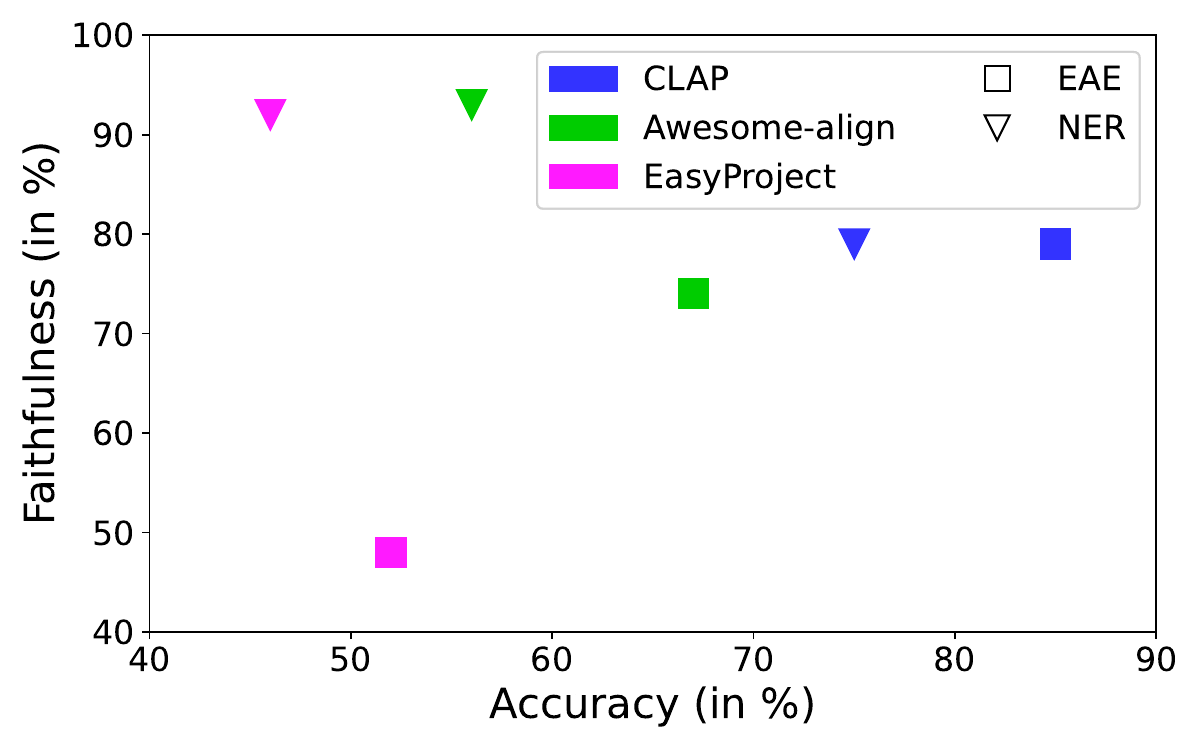}
    \caption{Reporting faithfulness and accuracy (in \%) for the different label projection models on EAE and NER. The closer the model is to the top-right, the better it is.}
    \label{fig:intrinsic-eval-plot}
\end{figure}

\begin{table}[t!]
    \small
    \centering
    \begin{tabular}{l|cc|c}
        \toprule
         & ar & zh & \textbf{Avg}\\ 
        \midrule
        LLM-Infer               & 16.9 & 24.0 & 20.5 \\
        \midrule
        Zero-shot$^*$           & 40.3 & 51.9 & 46.1 \\
        \midrule
        Awesome-align       & 48.6 & 54.5 & 51.6 \\
        EasyProject         & 38.5 & 56.3 & 47.4 \\
        \modelName{} (ours) & \textbf{49.3} & \textbf{58.6} & \textbf{54.0} \\
        \bottomrule
    \end{tabular}
    \caption{Extrinsic evaluation of the different label projection techniques regarding downstream model performance using translate-train and the LLM-Infer baseline for EAE. Avg = Average. $^*$ indicates the reproduced results of X-Gear~\cite{huang-etal-2022-multilingual-generative}.}
    \vspace{-0.5em}
    \label{tab:eae-gmt-results}
\end{table}

\begin{table*}[t!]
    \centering
    \small
    \setlength\tabcolsep{1.9pt}
    \begin{tabular}{l|cccccccccccccccccccc}
        \toprule
        \textbf{Lang} & \textbf{af} & \textbf{ar} & \textbf{bg} & \textbf{bn} & \textbf{de} & \textbf{el} & \textbf{es} & \textbf{et} & \textbf{eu} & \textbf{fa} & \textbf{fi} & \textbf{fr} & \textbf{he} & \textbf{hi} & \textbf{hu} & \textbf{id} & \textbf{it} & \textbf{ja} & \textbf{jv} & \textbf{ka} \\
        \midrule
        LLM-Infer & 50.9 & 24.8 & 66.9 & 12.0 & 44.2 & 42.2 & 59.5 & 41.6 & 36.7 & 19.5 & 46.7 & 53.5 & 15.6 & 18.9 & 20.6 & 30.3 & 56.0 & 35.7 & 28.7 & 21.7 \\
        \midrule
        Zero-shot & 77.4 & 48.1 & 82.8 & 77.0 & 78.8 & 80.6 & 74.5 & 78.7 & 61.4 & 69.2 & 79.3 & 79.4 & 57.3 & 70.6 & 80.8 & 53.1 & 79.4 & 19.1 & 58.5 & 72.3 \\
        \midrule
        Awesome-align & \textbf{77.9} & 46.0 & \textbf{81.0} & \textbf{81.2} & \textbf{78.8} & 71.7 & 65.3 & \textbf{78.0} & 66.8 & 46.4 & \textbf{77.4} & 78.2 & 55.3 & \textbf{73.9} & \textbf{77.4} & 52.8 & 79.3 & 20.3 & 56.3 & 70.4 \\
        EasyProject & 76.1 & 34.4 & \textbf{81.0} & 78.6 & \textbf{78.8} & 69.3 & 70.5 & 73.9 & 54.8 & 49.1 & \textbf{77.8} & 78.8 & \textbf{61.1} & 73.0 & 75.6 & 51.0 & 79.0 & 41.3 & 62.4 & 66.4 \\
        \modelName{} & 74.4 & \textbf{48.7} & \textbf{81.0} & 78.1 & 78.4 & \textbf{75.9} & \textbf{74.7} & 77.4 & \textbf{68.8} & \textbf{59.0} & 75.9 & \textbf{79.4} & 58.4 & 73.1 & 72.4 & \textbf{56.1} & \textbf{80.1} & \textbf{45.3} & \textbf{64.8} & \textbf{70.5} \\
        \midrule
        & \textbf{kk} & \textbf{ko} & \textbf{ml} & \textbf{mr} & \textbf{ms} & \textbf{my} & \textbf{nl} & \textbf{pt} & \textbf{ru} & \textbf{sw} & \textbf{ta} & \textbf{te} & \textbf{th} & \textbf{tl} & \textbf{tr} & \textbf{ur} & \textbf{vi} & \textbf{yo} & \textbf{zh} & \multicolumn{1}{|c}{\textbf{Avg}} \\
        \midrule
        LLM-Infer & 20.9 & 18.5 & 11.1 & 16.5 & 46.5 & 10.1 & 64.3 & 46.4 & 22.7 & 33.4 & 12.8 & 9.2 & 19.8 & 46.1 & 31.0 & 11.6 & 37.3 & 28.6 & 41.0 & \multicolumn{1}{|c}{32.1} \\
        \midrule
        Zero-shot & 51.9 & 57.5 & 66.4 & 65.3 & 53.4 & 65.8 & 83.0 & 80.0 & 74.2 & 68.4 & 60.3 & 62.1 & 0.4 & 74.5 & 65.6 & 62.2 & 75.0 & 34.1 & 24.6 & \multicolumn{1}{|c}{64.2} \\
        \midrule
        Awesome-align & \textbf{47.7} & 57.7 & \textbf{63.4} & \textbf{62.4} & 70.7 & 54.1 & \textbf{83.0} & 75.8 & 64.8 & 70.1 & \textbf{62.4} & \textbf{55.4} & 2.4 & \textbf{80.9} & 62.8 & 53.7 & 66.4 & \textbf{61.5} & 45.4 & \multicolumn{1}{|c}{63.5} \\
        EasyProject & 31.7 & 48.2 & 56.5 & 59.8 & 71.7 & 60.3 & 81.9 & \textbf{79.6} & 66.3 & \textbf{71.5} & 53.2 & 54.2 & 11.4 & 78.2 & \textbf{66.8} & \textbf{63.8} & 65.6 & 68.8 & 42.0 & \multicolumn{1}{|c}{63.2} \\
        \modelName{} & 42.8 & \textbf{60.1} & 60.3 & 61.4 & \textbf{73.5} & \textbf{61.5} & 82.2 & 78.2 & \textbf{68.3} & 70.6 & 59.6 & 53.1 & \textbf{13.2} & 74.6 & 62.9 & 32.9 & \textbf{75.8} & 59.6 & \textbf{49.7} & \multicolumn{1}{|c}{\textbf{64.9}} \\
        \bottomrule
    \end{tabular}
    \caption{Extrinsic evaluation of the different label projection techniques in terms of downstream model performance using translate-train and the LLM-Infer baseline for NER. Avg = Average.}
    \label{tab:ner-results}
\end{table*}

\subsection{Intrinsic Evaluation}
\label{sec:intrinsic-eval}
We first evaluate \modelName{} by directly evaluating the label projection quality, mainly focusing on evaluating the \textbf{accuracy} and \textbf{faithfulness} of the translated labels, with the definition stated in \S~\ref{subsec:label_project_definition}.

We employ native speakers to assess the accuracy of label translations. The evaluation is carried out using a ranking framework, in which the label translations from each model are ranked, including the option for ties.
The final accuracy score represents the average percentage at which the model outperformed all other competitors.
We conduct this evaluation on 50 data samples for Chinese, Arabic, Hindi, and Spanish, respectively.

Faithfulness measures the fulfillment of the label projection constraint.
It is measured as a percentage of projected data points when all the translated labels are present in the translated input sentence $\big( y^{tgt}_{m} \in \textbf{x}^{tgt} \quad, \forall y^{tgt}_{m} \in \textbf{y}^{tgt} \big)$. The statistics use the complete test set on ACE and WikiANN.

\mypar{Results:}
The accuracy and faithfulness of the models are plotted together in Figure~\ref{fig:intrinsic-eval-plot}.
An ideal model should optimize both these metrics and thus, the closer the models are to the top-right, the better they are deemed.
Overall, this figure shows how \modelName{} performs the best intrinsically as it is the closest to the top-right for both the tasks.
For EAE, \modelName{} is better than all models in both the metrics, while for NER, \modelName{} compromises faithfulness slightly for stronger accuracy.
Awesome-align and EasyProject are both great at attaining higher projection rates but produce less accurate label translations.
Overall, intrinsic evaluation demonstrates that \modelName{} offers the optimal balance between accuracy and faithfulness on a qualitative basis.


\begin{table*}[ht!]
    \small
    \centering
    \setlength{\tabcolsep}{3.8pt}
    \begin{tabular}{p{4cm}ll|lp{3.5cm}l}
        \toprule
        \textbf{Source} & \textbf{Source} & \textbf{Target} & \multirow{2}{*}{\textbf{Technique}} & \textbf{Translated} & \multirow{2}{*}{\textbf{Explanation}} \\
        \textbf{Sentence} & \textbf{Label} & \textbf{Lang} & & \textbf{Label} \\
        \midrule
        \multirow{5}{4cm}{Born in Castelvetrano , Trapani and raised in Catania , he moved to Madrid to keep up his busy career .} & \multirow{5}{*}{Castelvetrano} & \multirow{5}{*}{hi} & \multirow{2}{*}{Awesome-align} & \hi{k\4-V\?lv\?\6{V}Ano} \hi{\6{V}ApAnF} & \multirow{2}{*}{Extra word} \\
        & & & & \textcolor{blue}{(Castelvetrano Trapani)} & \\ \cline{4-6}
        & & & EasyProject & Castelvetrano & No translation \\ \cline{4-6}
        & & & \multirow{2}{*}{\modelName} & \hi{k\4-V\?lv\?\6{V}Ano} & \multirow{2}{*}{Perfect} \\ 
        & & & & \textcolor{blue}{(Castelvetrano)} & \\
        \midrule
        \multirow{6}{4cm}{Unilaterally leading a coalition featuring tyrannies, effect such change remains a bad idea, Iraq's elections notwithstanding.} & \multirow{6}{*}{Iraq} & \multirow{6}{*}{zh} & \multirow{2}{*}{Awesome-align} & \zh{伊拉} & \multirow{2}{*}{Incomplete} \\
        & & & & \textcolor{blue}{(Ira-)} & \\ \cline{4-6}
        & & & \multirow{2}{*}{EasyProject} & \zh{尽管伊拉克} & \multirow{2}{*}{Extra word} \\ 
        & & & & \textcolor{blue}{(although Iraq)} & \\ \cline{4-6}
        & & & \multirow{2}{*}{\modelName} & \zh{伊拉克} & \multirow{2}{*}{Perfect} \\
        & & & & \textcolor{blue}{(Iraq)} & \\
        \bottomrule
    \end{tabular}
    \caption{Qualitative examples highlighting the error-cases of the baseline models along with explanations for Hindi (hi) and Chinese (zh). We also show how \modelName{} performs better and fixes the errors. \textcolor{blue}{Blue} text is English translation.}
    \label{tab:qual-study}
\end{table*}

\subsection{Extrinsic Evaluation}
\label{sec:extrinsic-eval}

Extrinsic evaluation implicitly assesses the effectiveness of various label projection methods in generating pseudo-training data for downstream tasks. The projected data is filtered based on the faithfulness constraint as $\mathcal{D}^{tgt}_{src}$ and used along with the original English data $\mathcal{D}_{src}$ for downstream training.

For \textbf{EAE}, we use X-Gear~\cite{huang-etal-2022-multilingual-generative}, the current state-of-the-art model for zero-shot cross-lingual EAE, as the downstream model.
For \textbf{NER}, we use $\text{XLM-RoBERTa}_{\text{large}}$ \cite{conneau-etal-2020-unsupervised} as our downstream model and follow XTREME \cite{DBLP:conf/icml/HuRSNFJ20} setup for implementations. All results are the average over five runs.

\mypar{Results:}
We present the EAE results in terms of argument classification F1 scores in Table~\ref{tab:eae-gmt-results}.
For reference, we also include the zero-shot baseline
(training only on $\mathcal{D}_{src}$).
Evidently, \modelName{} performs the best providing an average gain of 2.4 F1 points over the next best baseline of Awesome-align and a net gain of 7.9 F1 points over the zero-shot baseline.
This result is in sync with our intrinsic evaluation wherein \modelName{} performed the best for EAE.

The primary findings for the F1 scores of entity classification are shown in Table~\ref{tab:ner-results}. Overall, \modelName{} outperforms all benchmarks, achieving an absolute enhancement of 0.7 F1 points compared to the zero-shot baseline, and surpassing previous studies by 1.4-1.7 F1 points. The superior performance of the downstream model powered by \modelName{}, highlights \modelName{}'s efficacy in improving downstream tasks.

\mypar{LLM usage comparison - Direct Inference v/s Contextual Translation:}
We compare the fine-tuned models with \textbf{LLM-Infer}, a large language model (LLM) baseline directly inferring on the downstream task in the target language.
We utilize the chat version of Llama2-13B model \cite{DBLP:journals/corr/TouvronLlama23} for the baseline.~\footnote{Compared to the text version, the chat version of Llama2 provided better results.}
We explore various cross-lingual prompting strategies, following \citet{ahuja-etal-2023-mega} (complete experiments in Appendix~\ref{sec:llm-infer}), and report the performance for the best prompt here.
From results in Table~\ref{tab:eae-gmt-results} \& \ref{tab:ner-results}, we can assert how LLM-infer performs significantly poorer than any fine-tuned model, indicating how LLMs can't infer well on cross-lingual structured prediction.
On the other hand, we demonstrate that LLMs can be better utilized to do contextual translation, as used in \modelName{}, which leads to the best performance for both the downstream tasks.
Additional experiments with ChatGPT \cite{DBLP:journals/corr/abs-2005-14165} are also provided in Appendix~\ref{sec:llm-infer}.

\section{Analysis}


\subsection{Qualitative Analysis}
\label{sec:qual-eval}

Diving deeper, we qualitatively study typical error cases for the translated labels in four languages by different label projection techniques.
In $200$ examples of our study, we found that $18\%$ of the time, EasyProject predicts nothing due to markers dropped in the translated sentence, and for $19\%$, EasyProject simply copies the English label failing to translate it to the target language.
For Awesome-align, the majority of errors are due to additional words or incomplete label translations, similar to the observation presented in \cite{chen-etal-2023-frustratingly}.
This could be because it is hard for the word-alignment module to decide alignments between sub-words, leading to over-alignment or under-alignment.
We show two selected examples of our study from Hindi (hi) and Chinese (zh) in Table~\ref{tab:qual-study}, where we show how Awesome-align predicts extra words or incomplete words owing to misalignments, and EasyProject fails to translate the word for Hindi while producing extra tokens for Chinese.
In both cases, we show how \modelName{} makes accurate predictions and is more robust in maintaining accurate label translations.

\subsection{CLAP with Larger LLMs}
\label{sec:llm-clap}

We utilize a relatively small LLM Llama-2 \cite{DBLP:journals/corr/TouvronLlama23} with 13B parameters as $\mathcal{M}$ for our experiments with \modelName.
Here, we analyze the impact of utilizing a larger LLM for \modelName{}.
More specifically, we compare Llama-2-13B based \modelName{} with a larger GPT-3.5-Turbo \cite{DBLP:journals/corr/abs-2005-14165} based \modelName{} for five languages for EAE and NER in Table~\ref{tab:gpt-clap-results}.~\footnote{GPT-3.5-Turbo costs \$20-\$30 per language. Thus, owing to budget constraints, we restrict ourselves to 5 languages.}
We notice that using GPT-3.5-Turbo in \modelName{} is at par with the Llama-2 variant for medium to high-resource languages like Arabic (ar) and Chinese (zh).
On the other side, for lower-resourced languages like Yoruba (yo), Urdu (ur), and Kazakh (kk), GPT-3.5-Turbo introduces significantly larger improvements of 3 to 30 F1 points.
Thus, we hypothesize that larger multilingual LLMs can further improve \modelName{}, especially for low-resource languages, also evidenced in \citet{DBLP:journals/corr/abs-2308-16884}.

\begin{table}[t]
    \small
    \centering
    \resizebox{1.0\linewidth}{!}{
    \setlength{\tabcolsep}{2.5pt}
    \begin{tabular}{ll|cc|ccc}
        \toprule
        & \textbf{Model} & \multicolumn{2}{c|}{\textbf{EAE}} & \multicolumn{3}{c}{\textbf{NER}} \\
         & \textbf{Size} & ar & zh & yo & ur & kk \\ 
        \midrule
        \modelName{} (w/ Llama-2-13B) & 13B & \textbf{49.3} & \textbf{58.6} & 59.6 & 32.9 & 42.8 \\
        \modelName{} (w/ GPT-3.5-Turbo) & 175B & 49.1 & 58.4 & \textbf{62.3} & \textbf{60.1} & \textbf{46.6} \\
        \bottomrule
    \end{tabular}
    }
    \caption{Extrinsic evaluation of \modelName{} using Llama-2-13B and GPT-3.5-Turbo for five languages.}
    \label{tab:gpt-clap-results}
\end{table}

\subsection{Generalization to other translation models}
\label{sec:generalization-translation-models}

To verify the generalizability of our approach to other translation models, we perform an extrinsic evaluation of the label projection techniques on the EAE task using the mBART-50 many-to-many (MMT) \cite{kong-etal-2021-multilingual} translation model.
We show the results for this evaluation in Table~\ref{tab:eae-mmt-results}.
We see that \modelName{} performs the best with an average improvement of $2$ F1 points over the next best baseline of Awesome-align and $6.5$ F1 points over the zero-shot baseline.
This result shows our \modelName{} is a generalizable label projection technique and agnostic to the underlying translation model.


\begin{table}[t]
    \small
    \centering
    \begin{tabular}{l|cc|c}
        \toprule
         & ar & zh & \textbf{Avg} \\ 
        \midrule
        Zero-shot & 40.3 & 51.9 & 43.9 \\
        \midrule
        Awesome-align & 47.1 & 53.8 & 48.4 \\
        EasyProject & 36.5 & 55.6 & 45.4 \\
        \modelName{} (ours) & \textbf{48.2} & \textbf{56.9} & \textbf{50.4} \\
        \bottomrule
    \end{tabular}
    \caption{Extrinsic evaluation of the different label projection techniques using translate-train for EAE using the mBART-50 many-to-many translation model.}
    \label{tab:eae-mmt-results}
\end{table}

\subsection{Ablation Study for \modelName{}}
\label{sec:ablation-study}
To study the impact of using instruction-tuned models for \textit{contextual translation}, we conduct an ablation study comparing \modelName{} with the following baselines which put extra focus on accuracy or faithfulness for contextual machine translation:
(1) \textbf{Independent} translation uses the translation model $\mathcal{T}$ to independently (without any context of the input sentence) translate the source text labels to the target language (i.e. $\textbf{y}^{tgt} = \mathcal{T}(\textbf{y}^{src})$),
(2) \textbf{Constrained} translation which uses a decoding constraint to carry out the faithfulness requirements.
More specifically, during translation, it limits the generation vocabulary to the tokens in the translated sentence $x^{tgt}$.
We follow \citet{de-cao-etal-2022-multilingual, lu-etal-2022-summarization} for implementing these constraints.


\begin{table}[t]
    \small
    \centering
    \begin{tabular}{l|cc|c}
        \toprule
         & ar & zh & \textbf{Avg}\\ 
        \midrule
        Zero-shot & 40.3 & 51.9 & 43.9 \\
        \midrule
        Independent & 44.8 & 54.3 & 47.6 \\
        Constrained & 45.6 & 55.6 & 48.8 \\
        \modelName{} (ours) & \textbf{48.2} & \textbf{56.9} & \textbf{50.4} \\
        \midrule
        Supervised & 63.2 & 69.7 & 65.0 \\
        \bottomrule
    \end{tabular}
    \caption{Ablation study comparing different contextual translation techniques for label projection. Performance is measured by downstream EAE performance.}
    \label{tab:ablation-results}
\end{table}

We extrinsically evaluate the model performances of the techniques on the task of EAE using the MMT translation model~\footnote{Since decoding-time constraints for the Constrained model can't be applied to GMT} and show the results in Table~\ref{tab:ablation-results}.
The independent model compromises faithfulness while the constrained model sacrifices accuracy - but both models outperform the zero-shot baseline.
\modelName{} provides high accuracy and faithfulness and achieves the best performance improving by 1.6 to 2.8 F1 over the ablation baselines.





\subsection{\modelName{} for Translate-Test}
\label{sec:translate-test}

\begin{table}[t]
    \small
    \centering
    \setlength{\tabcolsep}{2.5pt}
    \begin{tabular}{l|cc|ccc|c}
        \toprule
        & \multicolumn{2}{c|}{\textbf{EAE}} & \multicolumn{3}{c|}{\textbf{NER}} & \textbf{Avg} \\
         & ar & zh & it & es & id \\ 
        \midrule
        Zero-shot & 36.3 & 47.3 & 79.4 & 74.5 & 53.1 & 58.1 \\
        \midrule
        Awesome-align & 32.8 & 30.1 & \textbf{77.5} & 69.6 & 51.4 & 52.3 \\
        EasyProject & 17.0 & 11.5 & 65.9 & 62.6 & 51.8 & 41.8 \\
        \modelName{} (ours) & \textbf{34.3} & \textbf{39.5} & 73.4 & \textbf{75.0} & \textbf{57.4} & \textbf{55.9} \\
        \bottomrule
    \end{tabular}
    \caption{Extrinsic evaluation of the different label projection techniques using translate-test using GMT for EAE and NER. Avg = Average}
    \label{tab:translate-test}
\end{table}

Another popular technique for cross-lingual transfer is translate-test \cite{DBLP:conf/icml/HuRSNFJ20, ruder-etal-2021-xtreme} which was discussed in \S~\ref{sec:translate-train-background}.
As part of this analysis, we study the applicability of \modelName{} for translate-test using extrinsic evaluation on Arabic (ar) and Chinese (zh) for EAE and Italian (it), Spanish (es), and Indonesian (id) for NER.
We show the results in Table~\ref{tab:translate-test}.
Overall, we see how \modelName{} outperforms both the other methods significantly achieving the best scores for $4$ out of the $5$ languages.
EasyProject performs the worst as it uses the translation model twice causing higher error propagation.
We also note how translate-test doesn't yield improvements over the zero-shot baseline, especially for EAE as it requires using label projection twice (once for trigger and once for arguments), thus leading to error propagation.

\section{\modelName{} for Low-Resource Languages}
\label{sec:low-resource}


\begin{table}[t]
    \small
    \centering
    \begin{tabular}{lrrrrr}
    \toprule
        \textbf{Lang} & \textbf{ha} & \textbf{ig} & \textbf{ny} & \textbf{rw} & \textbf{sn} \\
        \midrule
        Zero-shot & \textbf{72.9} & 46.4 & 49.0 & 45.0 & 50.2 \\
        \midrule
        Awesome-align & 72.2 & \textbf{64.1} & \textbf{64.9} & \textbf{55.9} & 55.4\\
        EasyProject & 72.0 & 54.6 & 50.5 & 54.5 & 42.5 \\
        \modelName{} (ours) & 69.9 & 60.5 & 58.7 & 53.6 & \textbf{59.7} \\
        \midrule
        & \textbf{sw} & \textbf{xh} & \textbf{yo} & \textbf{zu} & \textbf{qu} \\
        \midrule
        Zero-shot & \textbf{88.6} & 61.0 & \textbf{33.6} & \textbf{67.1} & 37.9 \\
        \midrule
        Awesome-align & 82.9 & 52.4 & 30.8 & 57.9 & 46.1 \\
        EasyProject & 81.3 & 50.6 & 25.2 & 44.3 & 44.1 \\
        \modelName{} (ours) & 80.7 & \textbf{61.3} & 30.6 & 54.4 & \textbf{48.7} \\
    \bottomrule
    \end{tabular}
    \caption{Extrinsic evaluation of the different label projection techniques using translate-train using GMT for NER for 10 low-resource languages.}
    \label{tab:low-resource}
\end{table}

To cater our model to a wide range of languages, we study the applicability of \modelName{} for low-resource languages.
Specifically, we consider the task of NER for 10 low-resource languages from Africa and South America.
For the test datasets, we utilize MasakhaNER \cite{adelani-etal-2022-masakhaner} for 9 African languages: Hausa (ha), Igbo (ig), Chichewa (ny), Kinyarwanda (rw), chShona (sn), Kiswahili (sw), isiXhosa (xh), Yorùbá (yo), isiZulu (zu), and refer to \citet{zevallos-etal-2022-introducing} for the South American language Quechua (qu).
We conduct extrinsic evaluation of translate-train models transferring from the English CoNLL training data\footnote{For qu, we only use 3,000 CoNLL training data points due to budget constraints.} using the GMT model and
present the results in Table~\ref{tab:low-resource}.
We observe that this is a particularly challenging setting as all the label projection techniques fail to improve over the zero-shot model for 4 languages.
Our model \modelName{} improves for 6 languages and performs the best for 3 languages.
This result is particularly encouraging as our model uses a small and English-centric 13B Llama-2 model and utilizing larger multilingual LLMs will amplify these improvements further (as shown in \S~\ref{sec:llm-clap}).~\footnote{Owing to budget constraints, we left the exploration as future work.}

\section{Related Works}

\paragraph{Zero-shot Cross-lingual Structure Extraction}
Since the emergence of strong multilingual models \cite{devlin-etal-2019-bert, conneau-etal-2020-unsupervised}, various works have focused on zero-shot cross-lingual learning \cite{DBLP:conf/icml/HuRSNFJ20, ruder-etal-2021-xtreme} and code-switching \cite{garg-etal-2018-code, hsu-etal-2023-code} for various structure extraction tasks like named entity recognition \cite{DBLP:journals/corr/abs-2101-11112, yang-etal-2022-crop}, relation extraction \cite{ni-florian-2019-neural, subburathinam-etal-2019-cross}, slot filling \cite{krishnan-etal-2021-multilingual}, and semantic parsing \cite{nicosia-etal-2021-translate-fill, sherborne-lapata-2022-zero}.
Recent works have focussed on building datasets \cite{pouran-ben-veyseh-etal-2022-mee, parekh-etal-2023-geneva}, benchmarking \cite{DBLP:journals/corr/abs-2311-09562} as well as developing novel modeling designs exploring the usage of parse trees \cite{subburathinam-etal-2019-cross, ahmad-etal-2021-syntax, hsu-etal-2023-ampere}, data projection \cite{yarmohammadi-etal-2021-everything}, pooling strategies \cite{DBLP:journals/corr/abs-2302-11365} and generative models \cite{hsu-etal-2022-degree, huang-etal-2022-multilingual-generative} to improve cross-lingual transfer.
We utilize the state-of-the-art model X-Gear \cite{huang-etal-2022-multilingual-generative} and XLM-R \cite{conneau-etal-2020-unsupervised} as the downstream models for EAE and NER respectively, and improve them further using \modelName-guided translate-train.

\paragraph{Label Projection Techniques}
Several works have attempted to solve label projection for various structure extraction tasks such as semantic role labeling \cite{aminian-etal-2017-transferring, fei-etal-2020-cross}, slot filling \cite{xu-etal-2020-end},
semantic parsing \cite{moradshahi-etal-2020-localizing, awasthi-etal-2023-bootstrapping}, 
NER \cite{ni-etal-2017-weakly, stengel-eskin-etal-2019-discriminative}, and question-answering \cite{lee-etal-2018-semi, lewis-etal-2020-mlqa, DBLP:conf/aaai/BorneaPRFS21}.
The earliest works \cite{yarowsky-etal-2001-inducing, akbik-etal-2015-generating} utilized statistical word-alignment techniques like GIZA++ \cite{och-ney-2003-systematic} or fast-align \cite{dyer-etal-2013-simple} for locating the labels in the translated sentence.
Recent works \cite{chen-etal-2023-frustratingly} have also explored the usage of neural word aligners like QA-align \cite{nagata-etal-2020-supervised} and Awesome-align \cite{dou-neubig-2021-word}.
Another set of works has explored the paradigm of mark-then-translate using special markers like quote characters ("") \cite{lewis-etal-2020-mlqa}, XML tags (<a>) \cite{DBLP:conf/icml/HuRSNFJ20}, and square braces ([0]) \cite{chen-etal-2023-frustratingly} to locate the translated labels.
Overall, both these techniques can be error-prone and have poorer translation quality \cite{akbik-etal-2015-generating}, as shown in \S~\ref{sec:intrinsic-eval} and ~\ref{sec:qual-eval}.
A recent concurrent work CODEC \cite{DBLP:journals/corr/abs-2402-03131} improves the translation quality of text with markers by constrained decoding and data augmentation.


\section{Conclusion and Future Work}

In our work, we propose a novel approach \modelName{} for label projection, which utilizes contextual machine translation using instruction-tuned language models.
Experiments on two structure prediction tasks of EAE and NER across 39 languages demonstrate the effectiveness of \modelName{} compared to other label projection techniques.
Intrinsic evaluation provides deeper insights that justify our model improvements.
Additional experiments using larger LLMs, various translation models, translate-test paradigm, and 10 extremely low-resource languages demonstrate the generalizability and future potential of \modelName{} for cross-lingual structured prediction.

\section*{Acknowledgements}
We thank Xueqing Wu, Yang Chen, Kareem Ahmed, Syed Shahriar, Tao Meng, and Sidi Lu for their valuable insights, experimental setups, intrinsic evaluation, paper reviews, and constructive comments.
We thank the anonymous reviewers for their feedback.
This work was partially supported by NSF 2200274, AFOSR MURI via Grant \#FA9550- 22-1-0380, Defense Advanced Research Project Agency (DARPA) grant \#HR00112290103/HR0011260656, and a Cisco Sponsored Research Award.

\section*{Limitations}
In our work, we show the effectiveness of our model \modelName{} on two representative structure prediction tasks of EAE and NER.
Its effectiveness for other structure prediction tasks remains unknown and can be extended in future works.
For \modelName{}, we utilized the 13B version of the Llama-2 model as the base instruction-tuned language model as a proof-of-concept for the effectiveness of \modelName{}.
Future works can explore the usage of other stronger LLMs to enhance the model performance.
Lastly, we would like to point out that our model doesn't improve over the zero-shot model for several languages, mainly owing to the limited language understanding and poor translation quality.
However, the focus of our work has been to show the effectiveness of our model with other used label projection techniques.
With growing model sizes and enhanced coverage of languages, we posit that our model will eventually be able to provide significant improvements for all languages.

\section*{Ethical Concerns}

We use an instruction-tuned language model (specifically LLama-2) as the base model for \modelName{}.
Since these instruction-tuned models are not trained equitably in all languages, the model generation quality may vary drastically for each language.
Furthermore, since these models are not trained on filtered safe content data, the model may potentially generate harmful content.

\bibliography{anthology,custom}
\bibliographystyle{acl_natbib}

\clearpage

\appendix

\section{Data Statistics}
\label{sec:appendix-data-statistics}

We present the extensive data statistics for the ACE and WikiANN datasets used for downstream model evaluation on EAE and NER respectively.
For ACE, we follow the pre-processing by \citet{huang-etal-2022-multilingual-generative} to retain 33 event types and 22 argument roles.
For WikiAnn, we follow the pre-processing steps described in \citet{rahimi-etal-2019-massively, DBLP:conf/icml/HuRSNFJ20}.
For ACE, Table~\ref{tab:ace-statistics} provides statistics about the number of events and arguments for each language.
For WikiANN, we present the statistics in Table~\ref{tab:wikiann-statistics}.

\begin{table}[ht]
    \small
    \centering
    \setlength{\tabcolsep}{4.5pt}
    \begin{tabular}{lcc|cc}
    \toprule
         & \textbf{Train} & \textbf{Dev} & \multicolumn{2}{c}{\textbf{Test}} \\
         Language & English & English & Arabic & Chinese \\
    \midrule
         \# Events & 4,202 & 450 & 198 & 190 \\
         \# Arguments & 4,859 & 605 & 287 & 336 \\
    \bottomrule
    \end{tabular}
    \caption{Data Statistics in terms of events and arguments of the ACE dataset for the downstream task of EAE. \# indicates `number of'.}
    \label{tab:ace-statistics}
\end{table}

\begin{table}[ht!]
    \centering
    \small
    \begin{tabular}{llrr}
        \toprule
        \textbf{Split} & \textbf{Language} & \textbf{\# Sentences} & \textbf{\# Entities} \\
        \midrule
        Train & English (en) & 20,000 & 27,931 \\
        \midrule
        Dev & English (en) & 10,000 & 14,146 \\
        \midrule
        \multirow{39}{*}{Test} & Afrikaans (af) & 1,000 & 1,487 \\
        & Arabic (ar) & 10,000 & 11,259 \\
        & Bulgarian (bg) & 10,000 & 14,060 \\
        & Bengali (bn) & 1,000 & 1,089 \\
        & German (de) & 10,000 & 13,868 \\
        & Greek (el) & 10,000 & 12,163 \\
        & Spanish (es) & 10,000 & 12,260 \\
        & Estonian (et) & 10,000 & 13,892 \\
        & Basque (eu) & 10,000 & 13,459 \\
        & Farsi (fa) & 10,000 & 10,742 \\
        & Finnish (fi) & 10,000 & 14,554 \\
        & French (fr) & 10,000 & 13,369 \\
        & Hebrew (he) & 10,000 & 13,698 \\
        & Hindi (hi) & 1,000 & 1,228 \\
        & Hungarian (hu) & 10,000 & 14,163 \\
        & Indonesian (id) & 10,000 & 11,447 \\
        & Italian (it) & 10,000 & 13,749 \\
        & Japanese (ja) & 10,000 & 13,446 \\
        & Javanese (jv) & 100 & 117 \\
        & Georgian (ka) & 10,000 & 13,057 \\
        & Kazakh (kk) & 1,000 & 1,115 \\
        & Korean (ko) & 10,000 & 14,423 \\
        & Malayalam (ml) & 1,000 & 1,204 \\
        & Marathi (mr) & 1,000 & 1,264 \\
        & Malay (ms) & 1,000 & 1,115 \\
        & Burmese (my) & 100 & 119 \\
        & Dutch (nl) & 10,000 & 13,725 \\
        & Portuguese (pt) & 10,000 & 12,823 \\
        & Russian (ru) & 10,000 & 12,177 \\
        & Swahili (sw) & 1,000 & 1,194 \\
        & Tamil (ta) & 1,000 & 1,241 \\
        & Telugu (te) & 1,000 & 1,171 \\
        & Thai (th) & 10,000 & 16,970 \\
        & Tagalog (tl) & 1,000 & 1,034 \\
        & Turkish (tr) & 10,000 & 13,587 \\
        & Urdu (ur) & 1,000 & 1,020 \\
        & Vietnamese (vi) & 10,000 & 11,305 \\
        & Yoruba (yo) & 100 & 111 \\
        & Chinese (zh) & 10,000 & 12,049 \\
        \bottomrule
    \end{tabular}
    \caption{Data Statistics in terms of sentences and entities of the WikiANN dataset for the downstream task of NER. \# indicates `number of'.}
    \label{tab:wikiann-statistics}
\end{table}

\section{Complete Results for Intrinsic Evaluation}

\subsection{Accuracy Evaluation}

\begin{figure*}[t]
    \centering
    \includegraphics[width=0.97\textwidth]{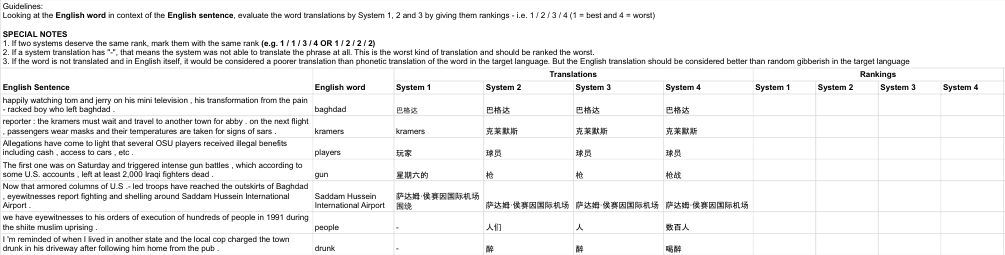}
    \caption{Annotation Interface for conducting the intrinsic evaluation for Accuracy. The shown examples are for Chinese, while the study was done for Hindi, Spanish, and Arabic as well.}
    \label{fig:accuracy-eval-interface}
\end{figure*}

Accuracy evaluation is done by $5$ native bilingual speakers for Chinese, Arabic, Hindi, and Spanish by ranking the translation quality of the translated labels.
The native speakers were undergraduate and graduate students who were well-versed in their respective native languages.
We present the interface of the google sheets along with the instructions shown to the annotators for Chinese in Figure~\ref{fig:accuracy-eval-interface}.
Similarly, annotation was performed for the other languages as well.
We present the complete results as an A/B comparison of the different techniques in terms of their win rates (i.e. percentage when A is better than B) in Table~\ref{tab:accuracy-results}.
We note how \modelName{} is more accurate than previous baselines of Awesome-align and EasyProject while at par with the Independent baseline.

\begin{table*}[ht]
    \small
    \centering
    \setlength{\tabcolsep}{4pt}
    \renewcommand{\arraystretch}{1.2}
    \begin{tabular}{lll|ccc|ccc|ccc|ccc}
        \toprule
        \multirow{2}{*}{\textbf{System 1}} & \multirow{2}{*}{\textbf{v/s}} & \multirow{2}{*}{\textbf{System 2}} & \multicolumn{3}{|c|}{\textbf{Arabic}} & \multicolumn{3}{c|}{\textbf{Chinese}} & \multicolumn{3}{c|}{\textbf{Hindi}} & \multicolumn{3}{c}{\textbf{Spanish}} \\ \cline{4-15}
         & & & \textbf{S1} & \textbf{Tie} & \textbf{S2} & \textbf{S1} & \textbf{Tie} & \textbf{S2} & \textbf{S1} & \textbf{Tie} & \textbf{S2} & \textbf{S1} & \textbf{Tie} & \textbf{S2} \\
        \midrule
        \modelName & & Awesome-align & \textbf{36\%} & 58\% & 6\% & \textbf{45\%} & 50\% & 5\% & \textbf{20\%} & 74\% & 6\% & \textbf{12\%} & 84\% & 4\% \\
        \modelName & & EasyProject & \textbf{52\%} & 32\% & 16\% & \textbf{56\%} & 39\% & 5\% & \textbf{42\%} & 48\% & 10\% & \textbf{30\%} & 66\% & 4\% \\
        \modelName & & Independent & 18\% & 60\% & \textbf{22\%} & 12\% & 71\% & \textbf{17\%} & \textbf{18\%} & 64\% & \textbf{18\%} & \textbf{24\%} & 68\% & 8\% \\
        Independent & & Awesome-align & \textbf{44\%} & 42\% & 14\% & \textbf{39\%} & 57\% & 4\% & \textbf{28\%} & 60\% & 12\% & \textbf{20\%} & 64\% & 16\% \\
        Independent & & EasyProject & \textbf{50\%} & 44\% & 6\% & \textbf{50\%} & 46\% & 4\% & \textbf{52\%} & 36\% & 12\% & \textbf{32\%} & 52\% & 16\% \\
        Awesome-align & & EasyProject & \textbf{42\%} & 26\% & 32\% & \textbf{34\%} & 50\% & 16\% & \textbf{42\%} & 42\% & 16\% & \textbf{26\%} & 64\% & 10\% \\
        \bottomrule
    \end{tabular}
    \caption{A/B comparison of the various label projection techniques for accuracy evaluation for the Google Translation model. Accuracy is measured as the label translation quality by native human speakers. Here, \textbf{S1} = System 1 is better, \textbf{S2} = System 2 is better, and \textbf{Tie} = similar quality. The better systems are highlighted in \textbf{bold}.}
    \label{tab:accuracy-results}
\end{table*}

\subsection{Faithfulness Evaluation}

We present the complete results for the faithfulness evaluation per language in Tables~\ref{tab:eae-faithfulness-results} and ~\ref{tab:wikiann-faithfulness-results} for EAE and NER tasks respectively.
For EAE, \modelName{} has the best faithfulness followed by Awesome-align.
For NER, Awesome-align and EasyProject have the highest faithfulness.

\begin{table}[t]
    \small
    \centering
    \setlength{\tabcolsep}{5pt}
    \begin{tabular}{l|cc|c}
        \toprule
        \textbf{Techniques} & \textbf{ar} & \textbf{zh} & \textbf{Avg.} \\
        \midrule
        Independent & 33 & 38 & 35\\
        Awesome-align & 66 & 83 & 74 \\
        EasyProject & 31 & 66 & 48 \\
        \modelName & 74 & 85 & \textbf{79} \\
        \bottomrule
    \end{tabular}
    \caption{Faithfulness evaluation of the various label projection techniques for EAE as a percentage of the times the translated labels were present in the translated input sentence. Numbers are in percentage (\%). Higher faithfulness is better and the best techniques are highlighted in \textbf{bold}.}
    \label{tab:eae-faithfulness-results}
\end{table}

\begin{table}[t]
    \small
    \centering
    \setlength{\tabcolsep}{4pt}
    \begin{tabular}{lccccccccc}
        \toprule
        \textbf{Techniques} & \textbf{af} & \textbf{ar} & \textbf{bg} & \textbf{bn} & \textbf{de} & \textbf{el} & \textbf{es} \\
        \midrule
        Independent & 78 & 66 & 67 & 74 & 79 & 57 & 70 \\
        Awesome-align & 99 & 95 & 98 & 92 & 99 & 98 & 99 \\
        EasyProject & 100 & 98 & 83 & 98 & 97 & 89 & 99 \\
        \modelName & 94 & 75 & 63 & 93 & 79 & 46 & 84 \\
        \midrule
        & \textbf{et} & \textbf{eu} & \textbf{fa} & \textbf{fi} & \textbf{fr} & \textbf{he} & \textbf{hi} \\
        \midrule
        Independent & 70 & 64 & 61 & 71 & 71 & 71 & 65 \\
        Awesome-align & 98 & 97 & 96 & 99 & 98 & 95 & 93 \\
        EasyProject & 97 & 94 & 99 & 98 & 99 & 94 & 36 \\
        \modelName & 92 & 91 & 72 & 92 & 74 & 80 & 90 \\
        \midrule
        & \textbf{hu} & \textbf{id} & \textbf{it} & \textbf{ja} & \textbf{jv} & \textbf{ka} & \textbf{kk} \\
        \midrule
        Independent & 68 & 77 & 74 & 68 & 66 & 64 & 56 \\
        Awesome-align & 98 & 99 & 99 & 58 & 98 & 95 & 94 \\
        EasyProject & 97 & 99 & 98 & 95 & 94 & 99 & 77 \\
        \modelName & 93 & 84 & 78 & 67 & 53 & 70 & 85 \\
        \midrule
        & \textbf{ko} & \textbf{ml} & \textbf{mr} & \textbf{ms} & \textbf{my} & \textbf{nl} & \textbf{pt} \\
        \midrule
        Independent & 63 & 57 & 73 & 80 & 53 & 76 & 76 \\
        Awesome-align & 96 & 88 & 92 & 99 & 90 & 99 & 97 \\
        EasyProject & 93 & 87 & 73 & 98 & 62 & 100 & 99 \\
        \modelName & 64 & 88 & 95 & 82 & 55 & 85 & 89 \\
        \midrule
        & \textbf{ru} & \textbf{sw} & \textbf{ta} & \textbf{te} & \textbf{th} & \textbf{tl} & \textbf{tr} \\
        \midrule
        Independent & 59 & 79 & 72 & 76 & 66 & 81 & 76 \\
        Awesome-align & 97 & 96 & 91 & 91 & 51 & 99 & 98 \\
        EasyProject & 99 & 97 & 91 & 87 & 99 & 99 & 98 \\
        \modelName & 66 & 94 & 96 & 90 & 57 & 58 & 94 \\
        \midrule
        & \textbf{vi} & \textbf{ur} & \textbf{yo} & \textbf{zh} & \textbf{Avg.} \\
        \midrule
        Independent & 74 & 74 & 45 & 66 & 69 \\
        Awesome-align & 83 & 97 & 92 & 92 & \textbf{93} \\
        EasyProject & 98 & 94 & 77 & 92 & 92 \\
        \modelName & 89 & 91 & 88 & 60 & 79 \\
        \bottomrule
    \end{tabular}
    \caption{Faithfulness evaluation of the various label projection techniques for NER as a percentage of the times the translated labels were present in the translated input sentence. Numbers are in percentage (\%). Higher faithfulness is better and the best techniques are highlighted in \textbf{bold}.}
    \label{tab:wikiann-faithfulness-results}
\end{table}

\section{Additional Implementation Details}
\label{sec:appendix-implementation-details}

\subsection{X-Gear}

X-Gear is used as the downstream model for EAE for extrinsic evaluation of the label projection techniques.
The original X-Gear work \cite{huang-etal-2022-multilingual-generative} explored two base multilingual models: mBART-50-large (mBART) \cite{kong-etal-2021-multilingual} and the mT5-base (mT5) \cite{xue-etal-2021-mt5}.
They also explored the usage of copy mechanism \cite{see-etal-2017-get} to prompt the models to predict strings from the input sentence.
In our work, we utilized mBART without copy (mBART), mT5 without copy (mT5), and mT5 with copy mechanism (mT5+Copy) as the downstream models.
We present details about the hyperparameter settings for these models in Table~\ref{tab:hyper-xgear}.
We run experiments for \modelName{} on a NVIDIA GeForce RTX 2080 Ti machine with support for 8 GPUs.

\subsection{XLM-R}

XLM-R \cite{conneau-etal-2020-unsupervised} is used as the downstream model for NER for extrinsic evaluation of the label projection techniques.
We mainly follow the XTREME \cite{DBLP:conf/icml/HuRSNFJ20} framework for setting up the task and model.
We present details about the hyperparameter settings for this model in Table~\ref{tab:hyper-xlmr}.
We run experiments for \modelName{} on a NVIDIA GeForce RTX 2080 Ti machine with support for 8 GPUs.

\begin{table}[h]
    \centering
    \small
    \begin{tabular}{lr}
        \toprule
        Base Model & XLM - Roberta - Large \\
        \# Training Epochs & 5 \\
        Training Batch Size & 32 \\
        Evaluation Batch Size & 32 \\
        Learning Rate & $2 \times 10^{-5}$ \\
        Weight Decay & 0 \\
        Max Sequence Length & 128 \\
        \# Accumulation Steps & 1 \\
        \# Saving Steps & 1000 \\
        \bottomrule
    \end{tabular}
    \caption{Hyperparameter details for the NER downstream XLM-R model.}
    \label{tab:hyper-xlmr}
\end{table}

\begin{figure}
    \centering
    \includegraphics[width=0.48\textwidth]{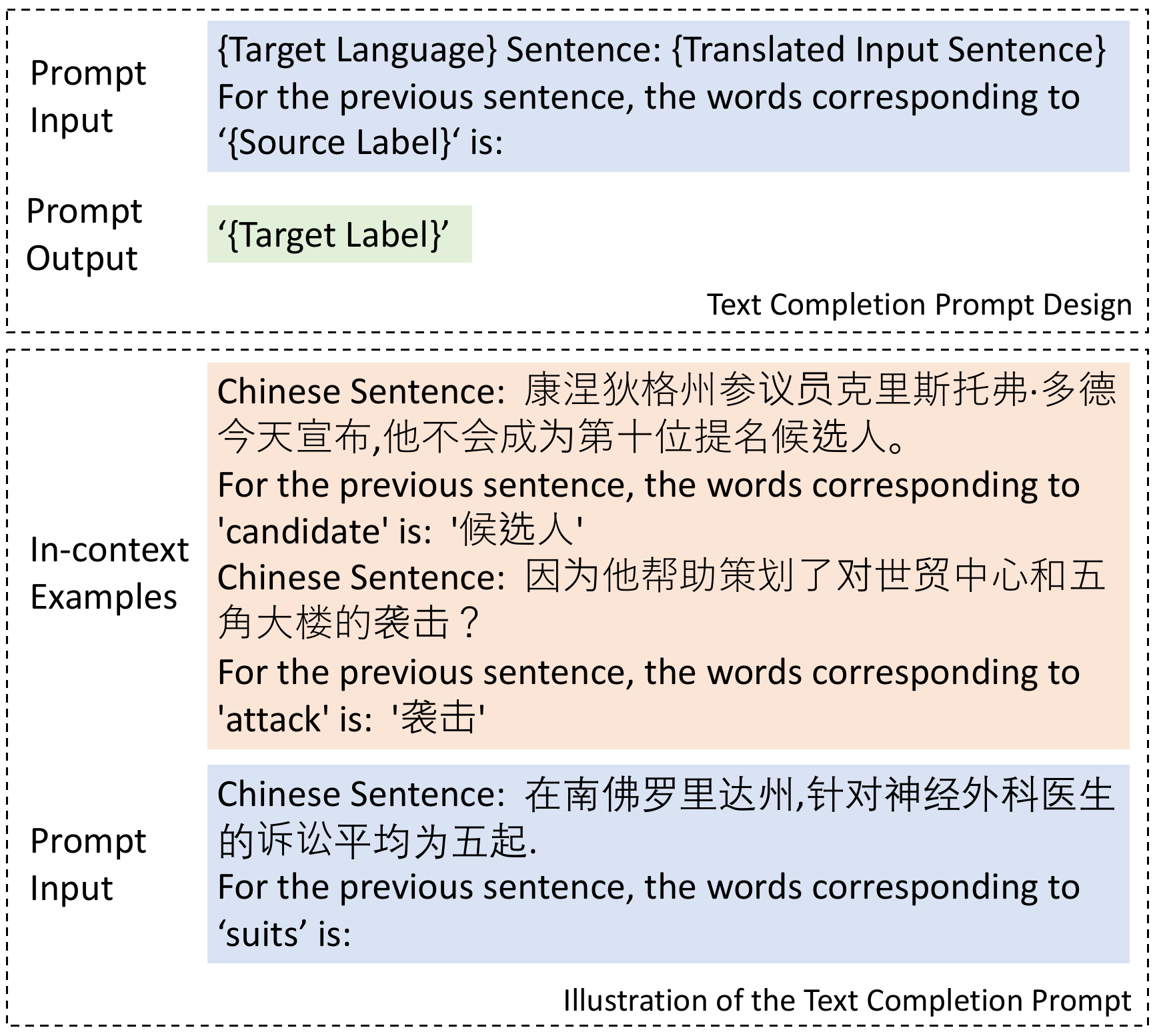}
    \caption{Illustration of the text-completion prompt used for contextual machine translation for our \modelName{} model.}
    \label{fig:clap-prompts}
\end{figure}

\begin{figure}
    \centering
    \includegraphics[width=0.48\textwidth]{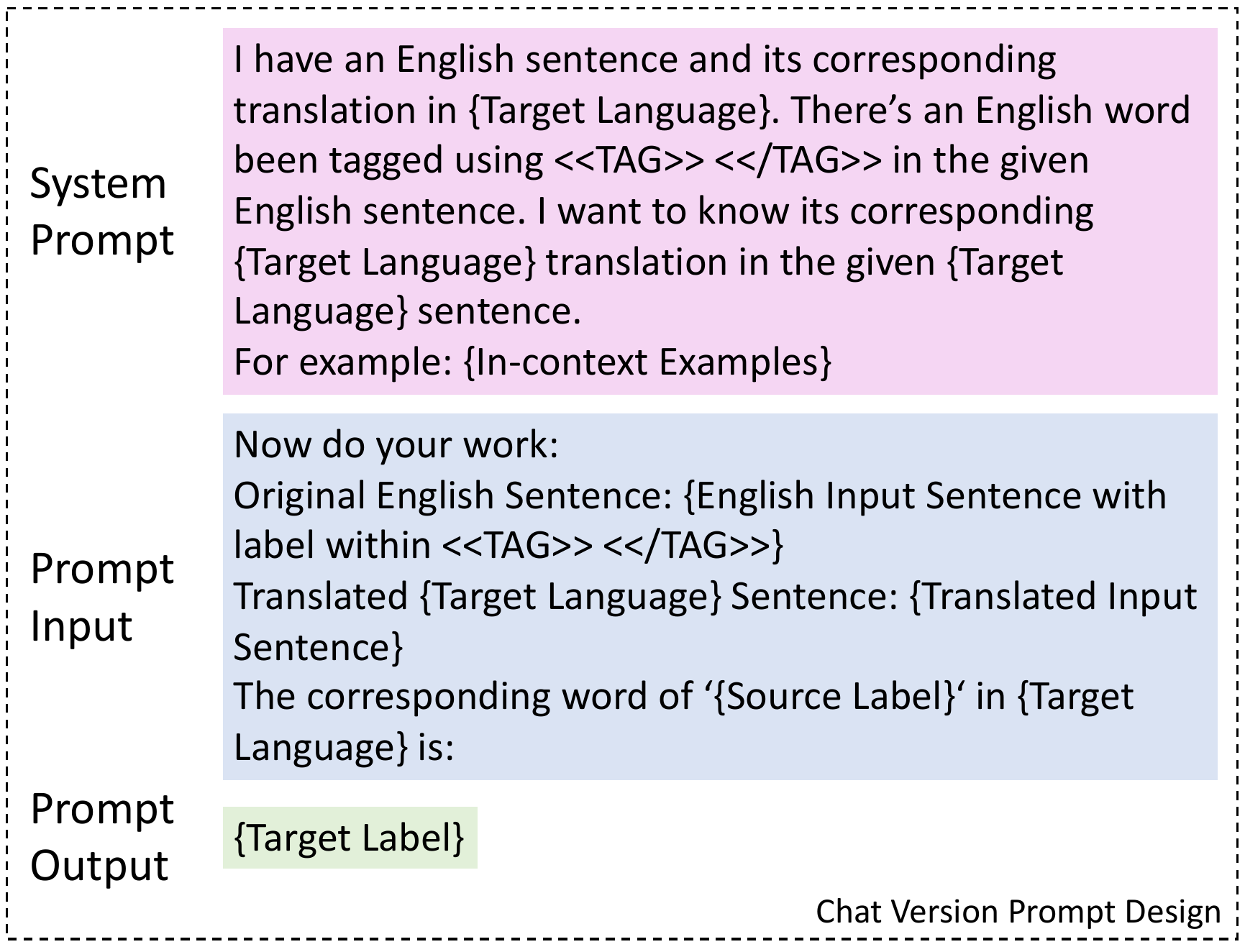}
    \caption{Illustration of the chat version prompt used for contextual machine translation for our \modelName{} model.}
    \label{fig:clap-gpt-prompts}
\end{figure}

\subsection{\modelName{}}

We provide a couple of prompt designs we used for our model in Figure~\ref{fig:clap-prompts} along with an illustration for Chinese.
We additionally provide a similar template for chat version of the model (which is used for experiments with GPT3.5-turbo as reported in \S~\ref{sec:llm-clap}) in Figure~\ref{fig:clap-gpt-prompts}.
We report the hyperparameter settings for our model in Table~\ref{tab:hyper-clap}.
We run experiments for \modelName{} on a NVIDIA GeForce RTX 2080 Ti machine with support for 8 GPUs.

\begin{table*}[ht]
    \small
    \centering
    \begin{tabular}{lrrr}
        \toprule
        & \textbf{mBART} & \textbf{mT5} & \textbf{mT5+Copy} \\
        \midrule
        Base Model & multilingual BART-Large & multilingual T5-Large & multilingual T5-Large \\
        Usage of copy & No & No & Yes \\
        Training Batch Size & 16 & 16 & 16 \\
        Eval Batch Size & 32 & 32 & 32 \\
        Learning Rate & $2 \times 10^{-5}$ & $1 \times 10^{-4}$ & $2 \times 10^{-5}$ \\
        Weight Decay & $1 \times 10^{-5}$ & $1 \times 10^{-5}$ & $1 \times 10^{-5}$ \\
        \# Warmup Epochs & 5 & 5 & 5 \\
        Gradient Clipping & 5 & 5 & 5 \\
        Max Training Epochs & 60 & 60 & 60 \\
        \# Accumulation Steps & 1 & 1 & 1 \\
        Beam Size & 4 & 4 & 4 \\
        Max Sequence Length & 350 & 350 & 350 \\
        Max Output Length & 100 & 100 & 100 \\
        \bottomrule
    \end{tabular}
    \caption{Hyperparameter details for the EAE downstream X-Gear model.}
    \label{tab:hyper-xgear}
\end{table*}

\begin{table}[h]
    \centering
    \small
    \begin{tabular}{lr}
        \toprule
        Base Model & llama-2-13b \\
        Temperature & 0.6 \\
        Top-p & 0.9 \\
        Maximum Generation Length & 64-128 \\
        \# In-context examples & 2 \\
        \bottomrule
    \end{tabular}
    \caption{Hyperparameter details for the \modelName{} model.}
    \label{tab:hyper-clap}
\end{table}

\subsection{EasyProject}
\label{sec:easyproject-implementation}

Compared to the original EasyProject work, we made certain changes in the re-implementation for our work to provide a fair comparison.
First, we use square-indexed markers (e.g. [0] and [/0]) compared to XML markers (e.g. <LOC> and </LOC>) used by EasyProject.
This is mainly because we obtained much higher retention rates using square-indexed markers (88.2\%) compared to XML markers (6.2\%) in our initial studies.
Secondly, the original EasyProject model uses a finetuned NLLB-200-3.3B model as the translation model.
Since we don't finetune \modelName{} or Awesome-align, we use the non-finetuned Google Machine Translation (GMT) model as the translation model.

\section{Complete results for Extrinsic Evaluation}

\subsection{Event Argument Extraction}

Here, we explore three versions of the X-Gear \cite{huang-etal-2022-multilingual-generative} model: mBART without copy (mBART), mT5 without copy (mT5), and mT5 with copy mechanism (mT5+Copy).
We present the extrinsic evaluation for EAE by training these three models with the label projection techniques for translate-train in Table~\ref{tab:eae-complete-gmt-results}.
Results indicate how \modelName{} performs the best across all the three variations of the model.

\begin{table}[t!]
    \small
    \centering
    \setlength{\tabcolsep}{2.5pt}
    \begin{tabular}{l|cc|cc|cc|c}
        \toprule
        & \multicolumn{2}{c|}{\textbf{mBART}} & \multicolumn{2}{c|}{\textbf{mT5}} & \multicolumn{2}{c|}{\textbf{mT5+Copy}} & \textbf{Avg} \\
         & ar & zh & ar & zh & ar & zh \\ 
        \midrule
        LLM-Infer               & - & - & - & - & 16.9$^+$ & 24.0$^+$ & 20.5 \\
        \midrule
        Zero-shot$^*$           & 36.3 & 47.3 & 36.7 & 51.0 & 40.3 & 51.9 & 43.9 \\
        \midrule
        Awesome-align       & 45.2 & 49.4 & \textbf{46.8} & 53.7 & 48.6 & 54.5 & 49.7 \\
        EasyProject         & 37.9 & 52.3 & 34.5 & 54.6 & 38.5 & 56.3 & 45.7 \\
        \modelName{} (ours) & \textbf{46.0} & \textbf{53.4} & 44.3 & \textbf{56.5} & \textbf{49.3} & \textbf{58.6} & \textbf{51.4} \\
        \bottomrule
    \end{tabular}
    \caption{Extrinsic evaluation of the different label projection techniques regarding downstream model performance using translate-train for EAE. Avg = Average. $^*$ indicates the reproduced results of X-Gear~\cite{huang-etal-2022-multilingual-generative}. Results for LLM-Infer (marked with $^+$) are independent of the XGear base model.}
    \label{tab:eae-complete-gmt-results}
\end{table}

\section{Large Language Model Direct Inference Analysis}
\label{sec:llm-infer}


Large language models (LLMs) have shown great zero-shot and few-shot capabilities for several tasks like sentiment analysis, machine translation, and question-answering \cite{DBLP:journals/corr/GuoHow23, DBLP:journals/corr/JiaoChatGPT23}.
However, employing a directly prompted LLM for information extraction and structured prediction tasks in cross-lingual settings is an under-studied area.
Current evidence, including recent studies by \citet{DBLP:journals/corr/HanInfo23} and \citet{DBLP:journals/corr/LiEvaluating2023}, indicates that LLM performance for these tasks, even for English, lags behind best fine-tuned models.
To this end in our work, we evaluate LLMs for direct inference on non-English structured prediction through our baseline \textbf{LLM-Infer}.

\begin{figure}
    \centering
    \includegraphics[width=0.48\textwidth]{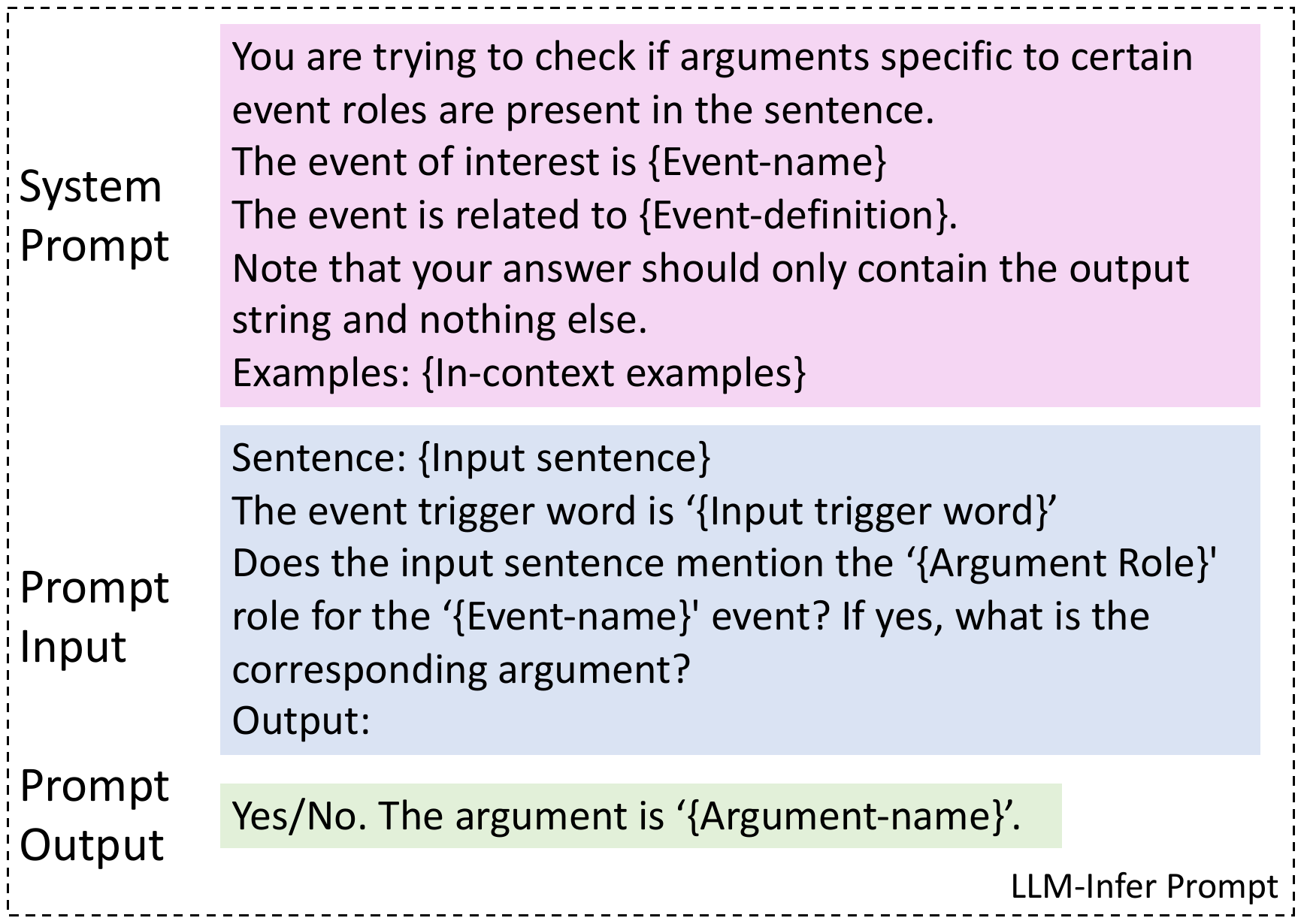}
    \caption{Illustration of the prompt used for the LLM-infer baseline to directly utilize LLMs for downstream structured prediction tasks.}
    \label{fig:gpt-infer-prompts}
\end{figure}

\begin{figure}
    \centering
    \includegraphics[width=0.48\textwidth]{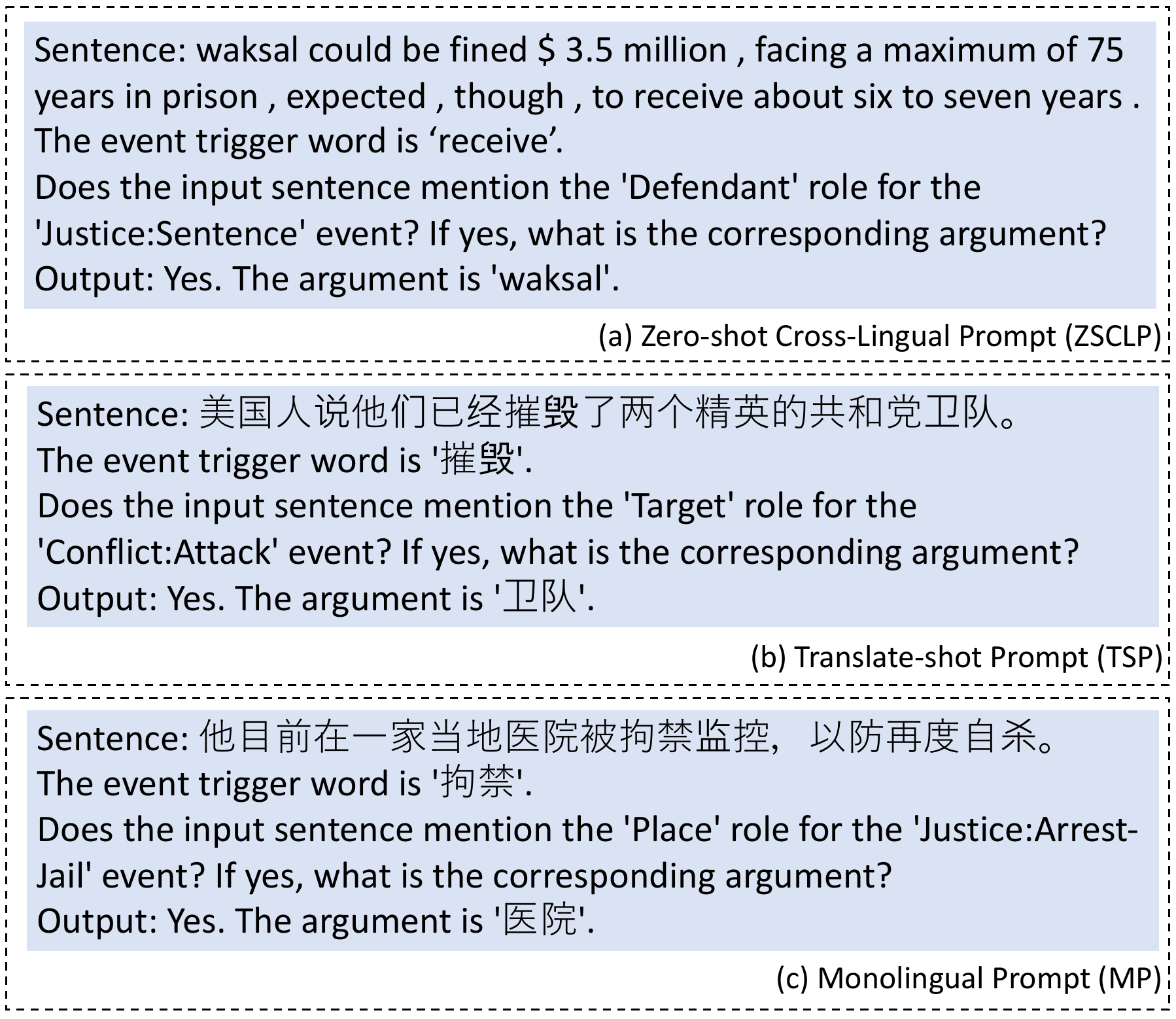}
    \caption{Illustration of the in-context examples used for the three different prompting strategies for LLM-Infer baseline.}
    \label{fig:gpt-in-context-examples}
\end{figure}

\begin{table*}[ht]
    \centering
    \small
    \setlength{\tabcolsep}{3.5pt}
    \begin{tabular}{lll|rr|rrr|r}
        \toprule
        \textbf{Base} & \textbf{Prompting} & \textbf{k-shot} & \multicolumn{2}{c|}{\textbf{EAE}} & \multicolumn{3}{c|}{\textbf{NER}} & \textbf{Avg} \\
        \textbf{Model} & \textbf{Strategy} &  & \textbf{ar} & \textbf{zh} & \textbf{hi} & \textbf{ms} & \textbf{yo} & \\
        \midrule
        Llama2-13b-chat & ZSCLP & 2 & 13.4 & 20.0 & 21.7 & 30.1 & 26.4 & 22.3 \\
        Llama2-13b-chat & ZSCLP & 4 & 14.2 & 17.9 & 39.5 & 38.3 & 31.9 & 28.4 \\
        Llama2-13b-chat & TSP & 2 & 16.9 & 24.0 & 18.9 & 46.5 & 28.6 & 27.0 \\
        Llama2-13b-chat & TSP & 4 & 8.7 & 22.8 & 17.5 & 43.5 & 36.2 & 25.7 \\
        Llama2-13b-chat & MP & 2 & 18.9 & 28.1 & 13.7 & 49.2 & 17.6 & 25.5 \\
        Llama2-13b-chat & MP & 4 & 11.9 & 26.0 & 13.7 & 61.5 & 17.4 & 26.1 \\
        \midrule
        GPT-3.5-turbo & ZSCLP & 2 & 15.8 & 22.3 & 64.4 & 50.7 & 39.7 & 38.6 \\
        GPT-3.5-turbo & ZSCLP & 4 & 15.9 & 23.6 & 65.0 & 53.0 & 39.0 & 39.3 \\
        GPT-3.5-turbo & TSP & 2 & 17.1 & 22.3 & 59.3 & 54.6 & 53.3 & 41.3 \\
        GPT-3.5-turbo & TSP & 4 & 17.2 & 24.5 & 52.3 & 57.2 & 48.8 & 40.0 \\
        GPT-3.5-turbo & MP & 2 & 15.3 & 25.2 & 59.5 & 64.1 & 51.0 & 44.7 \\
        GPT-3.5-turbo & MP & 4 & 19.5 & 28.8 & 58.5 & 65.4 & 48.5 & 44.1 \\
        \midrule
        Zero-shot Model & & & 40.3 & 51.9 & 70.6 & 53.4 & 34.1 & 50.1 \\
        \modelName{} Translate-train (Ours) & & & \textbf{49.3} & \textbf{58.6} & \textbf{73.1} & \textbf{73.5} & \textbf{59.6} & \textbf{62.8} \\
        \bottomrule
    \end{tabular}
    \caption{Evaluation of LLM-based inference and their comparison with our label projected translate-train model \modelName. This study is done on Event Argument Extraction (EAE) for two languages - Arabic (ar) and Chinese (zh) - and on Named Entity Recognition (NER) for three languages: Hindi (hi), Malay (ms), and Yoruba (yo).}
    \label{tab:llm-infer-results}
\end{table*}

We utilize two LLMs of varying sizes for LLM-Infer: Llama-2-chat (13B version) \cite{DBLP:journals/corr/TouvronLlama23} and GPT-3.5-Turbo \cite{DBLP:journals/corr/abs-2005-14165}.
We illustrate the prompts used for this baseline in Figure~\ref{fig:gpt-infer-prompts}.
Our LLM prompts involve 2-shot and 4-shot in-context examples, and we meticulously explore three distinct prompting strategies, specifically for the cross-lingual setting, following \citet{ahuja-etal-2023-mega} (also illustrated in Figure~\ref{fig:gpt-in-context-examples}).
These strategies are listed as follows: 

\begin{enumerate}
    \item \textbf{Zero-shot Cross-Lingual Prompt (ZSCLP)}: This strategy involves using k-shot examples from a pivot language (English in our study), which differs from the language of the test example, as shown in Figure~\ref{fig:gpt-in-context-examples}(a).
    \item \textbf{ Translate-shot Prompt (TSP)}: In this strategy, we first obtain k-shot examples from the pivot language and subsequently perform label projection (using \modelName) to the target language on these examples. These label-projected examples are used as in-context examples in the final prompt (Figure~\ref{fig:gpt-in-context-examples}(b)).
    \item \textbf{Monolingual Prompt (MP)}: This method uses k-shot human-labeled examples directly from the target language (Figure~\ref{fig:gpt-in-context-examples}(c)).
\end{enumerate}

While the first two strategies align with the zero-shot cross-lingual transfer setting, where the availability of data is limited to English, the third strategy offers a slight variation.
It presupposes the availability of a few examples in the target languages.
For a fair comparison, only the first two strategies are used to compare with \modelName, while the third strategy serves as a comparison datapoint for elucidating the difference between label-projected and human-labeled data as in-context examples.

We conduct this analysis on EAE across two languages and NER across three languages (as it's expensive to conduct this study for all the languages).
The selection of languages for NER is to consider both resource diversity (hi: medium-high resource; ms: medium resource; yo: low resource) and script diversity.
We compare these models with the zero-shot baseline and our proposed \modelName{} translate-train model.
We show the model performance results in terms of F1 scores for this study in Table~\ref{tab:llm-infer-results}.

This study reveals several insights:
(1) We observe that GPT-3.5-turbo significantly performs better than the Llama-2-13B model - signifying the importance of a larger model size.
(2) Comparing different prompting strategies, we observe little variation in model performance for the Llama-2-13B model, while a larger variation for GPT-3.5-turbo. 
Majorly, we observe that the label-projected in-context examples are better than the English examples, while human-labeled examples provide further gains of 3-4 F1 points.
(3) We observe that on average, the LLM-Infer models perform poorer than the zero-shot fine-tuned model.
These differences are massive for EAE, while for NER, LLM-Infer performs better for low-resource languages (ms and yo) using our label projected examples.
(4) Finally, we observe that \modelName{} performs the best across all tasks and all languages, even in cases where few-shot examples in target languages are used (MP prompting strategy).
All these insights validate \modelName{}'s manner of leveraging LLMs to solve zero-shot cross-lingual structured prediction tasks i.e. \modelName{} is better than direct LLM prompting.

\end{document}